\PassOptionsToPackage{numbers, compress}{natbib}

\documentclass{article}

\usepackage[preprint]{neurips_2026} 

\usepackage[utf8]{inputenc} 
\usepackage[T1]{fontenc}    
\usepackage{float}  
\usepackage{hyperref}       

\hypersetup{colorlinks=true, linkcolor=red!65!black, citecolor=blue!65!black, urlcolor=blue!65!black}

\usepackage{url}            
\usepackage{booktabs}       
\usepackage{amsfonts}       
\usepackage{nicefrac}       
\usepackage{microtype}      
\usepackage{xcolor}         

\usepackage{pifont}
\usepackage{colortbl}
\usepackage{graphicx}
\usepackage[export]{adjustbox}
\usepackage{multirow}
\usepackage{subcaption}
\usepackage{threeparttable}
\usepackage[percent]{overpic}  
\usepackage{makecell}
\usepackage{amsmath}
\usepackage{listings}
\usepackage{tcolorbox}
\usepackage{fancyvrb}
\usepackage{fvextra}
\usepackage{wrapfig}

\newcommand{\best}[1]{\textbf{#1}}
\DeclareSymbolFont{extraup}{U}{zavm}{m}{n}
\DeclareMathSymbol{\newcrossmark}{\mathalpha}{extraup}{129}
\definecolor{softgreen}{RGB}{46,160,67}
\definecolor{lightgreen}{RGB}{234,241,223}
\definecolor{lightred}{RGB}{252,234,220}
\definecolor{lightblue}{RGB}{216,230,231}
\definecolor{lightbrown}{RGB}{239,227,211}
\definecolor{lightpurple}{RGB}{198,213,234}
\definecolor{lightdark}{RGB}{223,218,212}

\usepackage[utf8]{inputenc}
\usepackage[T1]{fontenc}
\usepackage{microtype}
\usepackage{nicefrac}
\usepackage{xspace}

\usepackage{amsmath}
\usepackage{amsfonts}
\usepackage{amssymb}

\usepackage{booktabs}
\usepackage{enumitem}
\usepackage{xcolor}

\usepackage{graphicx}
\usepackage{subcaption}
\usepackage{wrapfig}

\usepackage{tikz}
\usetikzlibrary{shapes.geometric, arrows.meta, positioning, calc}

\usepackage{listings}
\definecolor{codeblue}{RGB}{0,0,255}
\definecolor{codered}{RGB}{200,0,0}
\definecolor{codegreen}{RGB}{0,128,0}
\definecolor{codegray}{RGB}{128,128,128}
\definecolor{codebg}{RGB}{248,248,248}
\lstdefinestyle{pythoncode}{
    language=Python,
    basicstyle=\small\ttfamily,
    keywordstyle=\color{codeblue},
    stringstyle=\color{codered},
    commentstyle=\color{codegreen}\itshape,
    numberstyle=\tiny\color{codegray},
    backgroundcolor=\color{codebg},
    frame=single,
    frameround=tttt,
    rulecolor=\color{black!20},
    breaklines=true,
    breakatwhitespace=true,
    showstringspaces=false,
    tabsize=4,
    xleftmargin=8pt,
    xrightmargin=8pt,
    aboveskip=10pt,
    belowskip=10pt,
    columns=flexible,
    keepspaces=true,
    escapeinside={(*@}{@*)},
}
\usepackage{tcolorbox}

\usepackage{algorithm}
\usepackage{algorithmic}

\usepackage{url}
\usepackage{hyperref}
\usepackage[capitalize,noabbrev]{cleveref}

\title{SQuID: A Benchmark for Quantitative Spatial Reasoning on Satellite Imagery}

\author{%
  Peter A.~Massih \\
  NEC Laboratories America \\
  \texttt{pmassih@nec-labs.com}
  \And
  Eric Cosatto \\
  NEC Laboratories America \\
  \texttt{cosatto@nec-labs.com}
  \And
  Martin Renqiang Min \\
  NEC Laboratories America \\
  \texttt{renqiang@nec-labs.com}
}

\begin{document}

\maketitle

\begin{abstract}
Vision-Language Models (VLMs) describe satellite images fluently but struggle with the quantitative spatial reasoning that remote sensing (RS) analysts perform routinely, such as counting buildings, measuring forest area in hectares and computing distances in meters. We introduce SQuID (Satellite Quantitative Intelligence Dataset), a benchmark of 2{,}000 questions over 587 images from four sources at 0.3 to 1.0\,m ground sampling distance, covering 24 question types in three difficulty tiers. Answers are computed from curated segmentation masks or, for 50 questions, from human consensus, and numeric answers are scored against tolerance ranges informed by human annotations. SQuID also releases the reference masks behind its answers, so that modular systems can be evaluated using reference rather than predicted masks, and 75{,}687 training question-answer pairs on disjoint images (SQuID-Train). Among 15 evaluated systems, the best zero-shot end-to-end VLM reaches 28.1\%, and RS-specialized VLMs score lower. As a baseline, we propose QASR (Quantitative Agentic Spatial Reasoning), in which an LLM writes a program that segments the image and measures the masks with geometric operators. QASR reaches 42.0\%. On the 1{,}950 mask-derived questions, running its programs on the reference masks instead of predicted masks raises accuracy from 42.2\% to 82.6\%, so segmentation is its main source of error. Fine-tuning Qwen2.5-VL-7B on SQuID-Train raises its accuracy from 16.1\% to 40.7\%, close to QASR. Multi-condition questions (Tier 3) remain the hardest in every setting, at 62.0\% for QASR with reference masks and 13.2\% after fine-tuning. These results leave clear room for end-to-end models that learn metric spatial reasoning, and SQuID provides the data to train and evaluate them.
\end{abstract}
\section{Introduction}
\label{sec:intro}

A remote sensing (RS) analyst who counts buildings or measures deforested area works directly on the image, tracing boundaries and measuring distances at a known ground sampling distance (GSD). Vision-Language Models (VLMs) instead pass the image through an encoder~\cite{radford2021learning, liu2023llava} that reduces it to a coarse grid of tokens. They can describe a forest well, yet they often cannot say how many hectares it covers.

These measurements are central to RS work, such as forest area for deforestation monitoring, solar farm capacity for energy planning and building counts for urban planning. Recent benchmarks show that VLMs fall short on them. On the Quantity category of RSVLM-QA the best evaluated model reaches only 37.8\%~\cite{zi2025rsvlmqabenchmarkdatasetremote}, and GPT-4V reaches only $R^2=0.25$ on tree counting~\cite{zhang2024goodcaptioningbadcounting}.

One plausible cause is how images are encoded. A vision encoder with $16{\times}16$ patches turns a $1024{\times}1024$ image into a $64{\times}64$ grid of tokens, and many VLMs merge tokens further, so metric quantities must be recovered from token features rather than measured on pixels. The connectors that project visual features into the language model can also change similarity relationships between image embeddings, with reported changes to 40 to 60\% of an image's nearest neighbors~\cite{li2025lostembeddingsinformationloss}. These are neighbors in embedding space, not locations within an image. Whether the needed information is lost in encoding or models have not learned to use it is still open. Studying this requires metric reference answers, a way to evaluate programs with reference masks, and training data to test what models can learn. SQuID provides all three, and we make the following contributions.

\textbf{SQuID.} A benchmark of 2,000 questions over 587 images from four sources at 0.3 to 1.0\,m GSD, with 24 question types grouped into basic quantification (Tier~1), spatial relationships (Tier~2) and complex multi-condition queries (Tier~3). Most answers are computed from curated segmentation masks, and numeric answers are scored against tolerance ranges informed by responses from 10 human annotators. Questions go beyond counting to areas, distances and percentages, for example ``What is the total area in hectares of urban land within 200\,m of forest?'' To our knowledge, no existing RS benchmark scores metric answers against tolerance ranges informed by human annotations.

\textbf{Reference masks and training data.} We release the reference masks used to compute the 1,950 mask-derived answers, so that modular systems can be evaluated using reference rather than predicted masks. SQuID-Train adds 75,687 question-answer pairs generated by the same pipeline on 1,443 images that do not overlap with the benchmark.

\textbf{QASR.} A modular baseline in which an LLM writes a Python program that calls segmentation models and geometric operators, so that areas, distances and counts are computed on pixel-aligned masks at the stated GSD. For ``How many buildings are within 200\,m of water?'', the program segments buildings and water, computes their distances and counts the buildings within 200\,m. QASR depends on segmentation quality. We assess this dependence by running the same programs on reference masks.

\textbf{Findings.} (i) The best evaluated zero-shot end-to-end VLM reaches 28.1\% overall and 10.8\% on Tier~3, and RS-specialized VLMs, including EarthDial~\cite{soni2025earthdial}, score lower overall. (ii) QASR reaches 42.0\%. Running its programs on the reference masks raises accuracy from 42.2\% to 82.6\% on the mask-derived questions, so segmentation is its main source of error. (iii) Program quality still matters. With reference masks, Tier~3 reaches only 62.0\%, and Llama-3.1-8B as program generator reaches 67.1\% on the same mask-derived questions. (iv) Fine-tuning Qwen2.5-VL-7B on SQuID-Train raises its accuracy from 16.1\% to 40.7\%, close to QASR, but only to 13.2\% on Tier~3. (v) On the 50 human-annotated questions, individual annotator responses score 62.6\% against the annotators' own consensus, while the 13 systems scored with the same rule reach 12 to 46\%.

\textbf{Data and Code Availability.} SQuID, including its reference masks and training split (SQuID-Train), is available at \url{https://huggingface.co/datasets/PeterAM4/SQuID}. The generation, calibration, scoring and QASR code is available at \url{https://github.com/PeterAMassih/SQuID}.
\section{Related Work}

\subsection{Quantitative Limitations of VLMs}

Recent benchmarks show that VLMs struggle with quantitative spatial queries. GEOBench-VLM~\cite{danish2025geobenchvlmbenchmarkingvisionlanguagemodels} evaluates 31 fine-grained geospatial tasks, and the best model in its reported evaluation, LLaVA-OneVision, reaches only 41.7\% accuracy on its multiple-choice questions. RSVLM-QA~\cite{zi2025rsvlmqabenchmarkdatasetremote} (162,373 question-answer pairs) finds its Quantity category the hardest, with the best evaluated model (Ovis2) at 37.8\%. A failure analysis~\cite{zhang2024goodcaptioningbadcounting} reports that GPT-4V reaches only about 8\% localization accuracy at an IoU threshold of 0.5 on DIOR-RSVG, and that its counts reach $R^2$ values of 0.68 for buildings, 0.61 for vehicles, 0.25 for trees and 0.08 for animals.

Specialized RS models show similar limits. VHM~\cite{pang2024vhmversatilehonestvision} reports a mean absolute error of 6.75 for counting despite 95\% accuracy on image modality recognition. RemoteCLIP~\cite{liu2024remoteclipvisionlanguagefoundation} is evaluated on scene classification, retrieval and object counting rather than on areas and distances in physical units. We evaluate three RS-specialized VLMs in \cref{sec:experiments}, GeoChat~\cite{kuckreja2023geochatgroundedlargevisionlanguage}, Geo-R1~\cite{georl2025} and EarthDial~\cite{soni2025earthdial}, the latter trained on 11.11M RS instruction pairs.

A study of contrastively trained VLMs also reports ``bag-of-words'' behavior, with some evaluated models performing at or below chance on relation and word-order tests~\cite{yuksekgonul2023visionlanguagemodelsbehavelike}. In one example, BLIP prefers ``the grass is eating the horse'' to ``the horse is eating the grass''. The authors link this behavior to the contrastive objective and retrieval data, which give models little incentive to encode word order, and show that composition-aware hard negatives improve compositional understanding without changing the architecture.

\subsection{Code Generation for Visual Reasoning}

Code generation offers an alternative in which measurements are computed explicitly instead of being estimated inside a model. ViperGPT~\cite{suris2023vipergptvisualinferencepython} generates Python programs that call vision APIs and reported state-of-the-art zero-shot results on compositional reasoning. CodeVQA~\cite{subramanian2023modularvisualquestionanswering} improves overall few-shot VQA accuracy by up to 5 percentage points over its baseline, with a gain of about 30 percentage points on GQA questions about the relative positions of objects. VisProg~\cite{gupta2022visualprogrammingcompositionalvisual} composes 20 modules through generated programs without task-specific training. These results support using explicit programs for spatial computations. None of these three systems targets RS imagery, where the GSD varies across sensors, land-cover classes are domain-specific and measurements must be expressed in physical units.

In RS, ThinkGeo~\cite{shabbir2025thinkgeo} evaluates tool selection, multi-step execution and final answers. Earth-Agent~\cite{feng2026earthagent} combines expert tools for RGB and spectral analysis. QASR focuses on geometric measurements computed from segmentation masks.

\subsection{Segmentation-grounded Vision-Language Models}

Another line of work grounds language in image regions through segmentation and bounding boxes. LISA~\cite{lai2024lisareasoningsegmentationlarge} decodes masks from the embedding of a dedicated segmentation token, which enables reasoning segmentation. GeoChat~\cite{kuckreja2023geochatgroundedlargevisionlanguage} adds task tokens for grounding and referring and outputs rotated bounding boxes with its text. Text2Seg~\cite{zhang2024text2segremotesensingimage} combines Grounding DINO, SAM and CLIP for text-guided segmentation without domain-specific training, and SegEarth-OV~\cite{li2024segearthov} upsamples encoder features to improve spatial detail.

More recent models bring pixel-level cues into end-to-end architectures. GeoPixel~\cite{shabbir2025geopixel} generates masks interleaved with its answers, and CoVT~\cite{qin2025chain} reasons over continuous visual tokens distilled from segmentation, depth, edge and self-supervised feature experts. These models use dense visual cues, but do not explicitly compute distances, buffered intersections or area-constrained counts through geometric operators. QASR instead computes these quantities on pixel-aligned masks. SQuID can evaluate both approaches, since it only scores the final answer.

\subsection{Positioning SQuID}

Existing benchmarks test different aspects of quantitative spatial reasoning. RSVQA~\cite{lobry2020rsvqa} includes count and area questions, but its high-resolution dataset uses five area bins, with 60.9\% of area answers in the 0\,m$^2$ bin. EarthVQA~\cite{wang2023earthvqaqueryableearthrelational} (208K question-answer pairs) focuses on relational VQA, HRVQA~\cite{li2023hrvqavisualquestionanswering} (1.07M pairs) covers ten question types including number and area, and GEOBench-VLM covers a broad range of RS tasks. SpatialVLM~\cite{chen2024spatialvlmendowingvisionlanguagemodels} trains models for spatial reasoning on natural rather than aerial images. EarthSpatialBench~\cite{xu2026earthspatialbench} evaluates distances, directions, topological relations and buffer-based queries, reporting mean absolute error and root mean squared error for numeric answers. We distinguish two regimes.

\noindent\textbf{Object counting} asks ``how many buildings, aircraft or vehicles?'' and focuses on identifying and enumerating instances without additional metric constraints. GEOBench-VLM tests this regime across several object categories.

\noindent\textbf{Quantitative metric spatial reasoning} asks ``what is the area of agricultural land within 50\,m of water?'' or ``for water bodies larger than 5\,ha, how much of their area lies within 200\,m of agricultural land?'' A program can answer these questions by segmenting land-cover classes, computing distance transforms, applying area and proximity filters and aggregating metric quantities.

SQuID targets mainly the second regime, scoring numeric answers against tolerance ranges informed by human annotations and releasing reference masks so that the same programs can be evaluated with predicted or reference masks.
\section{The SQuID Dataset}
\label{sec:dataset}

SQuID (Satellite Quantitative Intelligence Dataset) is an evaluation benchmark of 2,000 questions over 587 images from four sources at three GSDs (0.3, 0.5 and 1.0\,m). It targets counting, measuring and computing spatial relations in metric units. Of its answers, 1,577 are numeric and 423 are categorical (325 binary and 98 fragmentation questions). We keep the test set compact so that it stays fast to run, and release a much larger training split generated by the same pipeline (\cref{subsec:companion}).

\subsection{Dataset Construction}

We generate questions from three RS datasets with segmentation masks, DeepGlobe~\cite{Demir_2018}, EarthVQA~\cite{wang2023earthvqaqueryableearthrelational} and a photovoltaic dataset~\cite{essd-13-5389-2021}, and add 50 human-annotated questions on NAIP imagery~\cite{usgsnaip} (\cref{tab:sources}). EarthVQA publishes masks for its training and validation splits, and its benchmark images come from the validation split. DeepGlobe publishes masks only for its 803 training images, and the photovoltaic dataset has no official split, so for these two sources the benchmark images come from the labeled pool. No benchmark image appears in SQuID-Train.

\begin{table}[t]
\caption{SQuID sources. Answers are computed from reference masks, except for NAIP, where they come from human consensus.}
\label{tab:sources}
\centering
\small
\setlength{\tabcolsep}{4pt}
\begin{tabular}{@{}lcccrrl@{}}
\toprule
Source & GSD & Classes & Images & Questions & Answers & Terms \\
\midrule
DeepGlobe~\cite{Demir_2018} & 0.5\,m & 6 & 174 & 612 & masks & DeepGlobe terms \\
EarthVQA~\cite{wang2023earthvqaqueryableearthrelational} & 0.3\,m & 5 & 364 & 1,241 & masks & EarthVQA terms \\
Solar~\cite{essd-13-5389-2021} & 0.3\,m & 1 & 35 & 97 & masks & CC BY 4.0 \\
NAIP~\cite{usgsnaip} & 1.0\,m & 7 & 14 & 50 & human & Public domain \\
\midrule
Total & & & 587 & 2,000 & & \\
\bottomrule
\end{tabular}
\end{table}

The generation pipeline (1) extracts connected components from the masks with OpenCV contour detection, (2) computes geometric properties in metric units from the GSD, (3) performs spatial operations such as distance transforms, buffering and intersection, and (4) generates questions whose answers are computed from the mask geometry. A quota-based sampler controls the distribution across question categories, tiers and sources.

To calibrate tolerance ranges and obtain a human reference, 10 annotators answered the 50 NAIP questions, which cover 14 images at 1.0\,m GSD. They used a grid-based interface built on Turkle~\cite{turkle2023} to mark regions, give counts or select categorical answers (interface in the supplementary material). The annotators were researchers trained on the interface, not a separate cohort of RS experts. Over the 47 numeric questions, Krippendorff's $\alpha$ is 0.79~\cite{krippendorff2004reliability}. The right panel of \cref{fig:dataset_statistics} shows reliability by question type, with count questions highest ($\alpha = 0.959$) and proximity questions lowest ($\alpha = 0.424$).

\subsection{Evaluation Modes and Training Data}
\label{subsec:companion}

SQuID scores only the final answer, so any system can be evaluated. We define three evaluation modes (\cref{fig:eval_modes}). In the end-to-end mode, a system answers from the image, the question and the GSD. In the predicted-mask mode, a modular system runs its program on the masks of its own segmentation model. In the reference-mask mode, the same logged program is run again on the reference masks of the 1,950 mask-derived questions, with no new LLM call. Comparing the last two modes measures the effect of replacing the segmentation output while keeping the program fixed. We recommend that modular systems report both scores on the same questions.

\begin{figure}[t]
\centering
\resizebox{\linewidth}{!}{
\begin{tikzpicture}[
  font=\sffamily\footnotesize,
  box/.style={rectangle, rounded corners=2pt, draw, line width=0.6pt,
              minimum height=0.8cm, align=center, inner sep=2pt},
  inp/.style={box, fill=gray!10, draw=gray!60!black, minimum width=2.2cm, text width=2.1cm},
  mod/.style={box, fill=blue!8, draw=blue!55!black, minimum width=2.2cm, text width=2.1cm},
  e2e/.style={box, fill=orange!12, draw=orange!60!black, minimum width=5.8cm, text width=5.6cm},
  logged/.style={box, fill=blue!8, draw=blue!55!black, dashed, minimum width=4.8cm, text width=4.7cm},
  ref/.style={box, fill=green!12, draw=green!45!black, minimum width=4.8cm, text width=4.7cm},
  exe/.style={box, fill=blue!8, draw=blue!55!black, minimum width=2.2cm, text width=2.1cm, minimum height=1.8cm},
  arr/.style={-{Stealth[length=4.5pt]}, line width=0.7pt},
  elab/.style={font=\sffamily\scriptsize, inner sep=1.5pt, above},
  rowlab/.style={anchor=west, align=left, text width=2.6cm,
                 execute at begin node={\hyphenpenalty=10000\exhyphenpenalty=10000}},
  res/.style={anchor=west, align=left, text width=2.3cm,
              execute at begin node={\hyphenpenalty=10000\exhyphenpenalty=10000}}
]
\node[rowlab] at (0,0) {\textbf{(a) End-to-end}\\Tests perception and reasoning together.};
\node[inp] (ina) at (4.05,0) {Image, question\\and GSD};
\node[e2e] (vlm) at (8.45,0) {Vision-language model\\(encoder, connector, LLM)};
\draw[arr, orange!60!black] (ina) -- (vlm);
\draw[arr, orange!60!black] (vlm.east) -- node[elab]{answer} ++(1.1,0);
\node[res] at (12.55,0) {\textbf{28.1\%} best zero-shot\\[3pt]\textbf{40.7\%} trained on SQuID-Train};

\draw[gray!50, dashed] (0,-1.0) -- (14.85,-1.0);

\node[rowlab] at (0,-2.15) {\textbf{(b) Modular,\\predicted masks}\\Tests segmentation and program together.};
\node[inp] (qb) at (4.05,-1.7) {Question\\and GSD};
\node[mod] (llmb) at (6.65,-1.7) {LLM writes\\a program};
\node[inp] (imgb) at (4.05,-2.6) {Image};
\node[mod] (segb) at (6.65,-2.6) {Segmentation\\model};
\node[exe] (exeb) at (10.25,-2.15) {Program runs\\on the masks\\(area, distance,\\buffer, count)};
\draw[arr, blue!55!black] (qb) -- (llmb);
\draw[arr, blue!55!black] (imgb) -- (segb);
\draw[arr, blue!55!black] (llmb.east) -- node[elab]{program} (llmb.east -| exeb.west);
\draw[arr, blue!55!black] (segb.east) -- node[elab]{masks} (segb.east -| exeb.west);
\draw[arr, blue!55!black] (exeb.east) -- node[elab]{answer} ++(1.1,0);
\node[res] at (12.55,-2.15) {\textbf{42.0\%} QASR\\(GPT-5 + ConvNeXt)};

\draw[gray!50, dashed] (0,-3.3) -- (14.85,-3.3);

\node[rowlab] at (0,-4.45) {\textbf{(c) Modular,\\reference masks}\\Tests the program on reference masks.};
\node[logged] (progc) at (5.35,-4.0) {Same program, logged in (b)\\(no new LLM call)};
\node[ref] (gtc) at (5.35,-4.9) {Reference masks\\(replace predicted masks)};
\node[exe] (exec) at (10.25,-4.45) {Program runs\\on the masks\\(area, distance,\\buffer, count)};
\draw[arr, blue!55!black] (progc.east) -- node[elab]{program} (progc.east -| exec.west);
\draw[arr, green!40!black] (gtc.east) -- node[elab]{masks} (gtc.east -| exec.west);
\draw[arr, blue!55!black] (exec.east) -- node[elab]{answer} ++(1.1,0);
\node[res] at (12.55,-4.45) {\textbf{82.6\%}, up from 42.2\% on the same 1,950 questions\\[3pt]Tier~3 \textbf{62.0\%}};
\end{tikzpicture}}
\caption{\textbf{SQuID evaluation modes.} (a) An end-to-end system maps the image, question and GSD to an answer. (b) A modular system runs a generated program on predicted masks. (c) The same program is run again on reference masks, with no new LLM call. On the 1,950 mask-derived questions, this corrects 814 of the 1,127 QASR errors. With the same reference masks, Llama-3.1-8B programs reach 67.1\%, compared with 82.6\% for GPT-5. Scores in (a) and (b) use all 2,000 questions, while (c) uses the 1,950 questions with reference masks.}
\label{fig:eval_modes}
\end{figure}

\textbf{Reference masks.} For the 1,950 mask-derived questions, we release the masks from which each answer was computed. The 50 NAIP questions have human-consensus answers and no source masks.

\textbf{SQuID-Train.} Running the same pipeline, templates and sampler on 1,443 images that do not overlap with the benchmark yields SQuID-Train, 75,687 question-answer pairs for training end-to-end models and program generators. Model selection should use a held-out part of SQuID-Train, never the benchmark.

\textbf{Release.} We release the benchmark, reference masks, SQuID-Train, the 500 anonymized annotations and the generation, calibration and scoring code. Images and masks remain subject to their source terms (\cref{tab:sources}).

\subsection{Question Types and Difficulty Tiers}

\begin{figure}[t]
\centering
\begin{tabular}{cc}
\includegraphics[width=0.73\linewidth, valign=c]{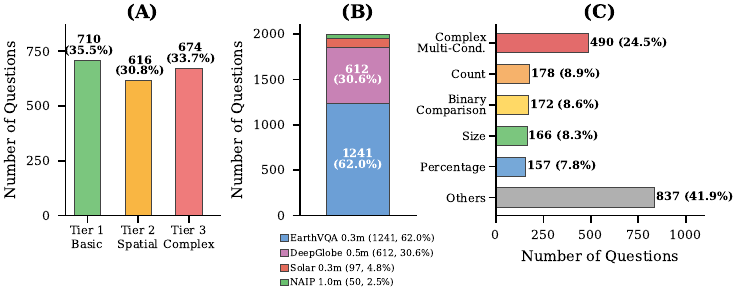}
\includegraphics[width=0.25\linewidth, valign=c]{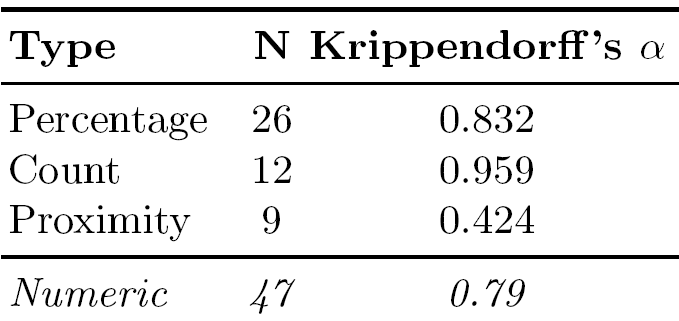}
\end{tabular}
\caption{\textbf{SQuID dataset composition.} \textbf{(A)} Three difficulty tiers. \textbf{(B)} Four data sources at 0.3 to 1.0\,m GSD. \textbf{(C)} Top 5 question types and the share of all other types. Right: inter-rater reliability on the 47 numeric questions, with 10 annotations per question. The three categorical questions are excluded from this table.}
\label{fig:dataset_statistics}
\end{figure}

SQuID groups its 24 question types into three tiers of 710, 616 and 674 questions (\cref{fig:dataset_statistics}).

\noindent\textbf{Tier 1: Basic Quantification.} Single-step quantities, such as coverage percentages, region counts, sizes, presence checks and coverage comparisons between two classes, with no multi-step spatial operation.

\noindent\textbf{Tier 2: Spatial Relationships.} Distances, proximity, connectivity, fragmentation and quantities derived from areas, such as solar capacity. Proximity questions involve explicit metric distances between land-cover classes.

\noindent\textbf{Tier 3: Complex Multi-Condition.} Questions that combine size thresholds with proximity constraints or area aggregation, such as total or average area.

\begin{figure}[t]
\centering
\begin{minipage}[t]{0.31\linewidth}
\centering
\includegraphics[width=0.72\linewidth]{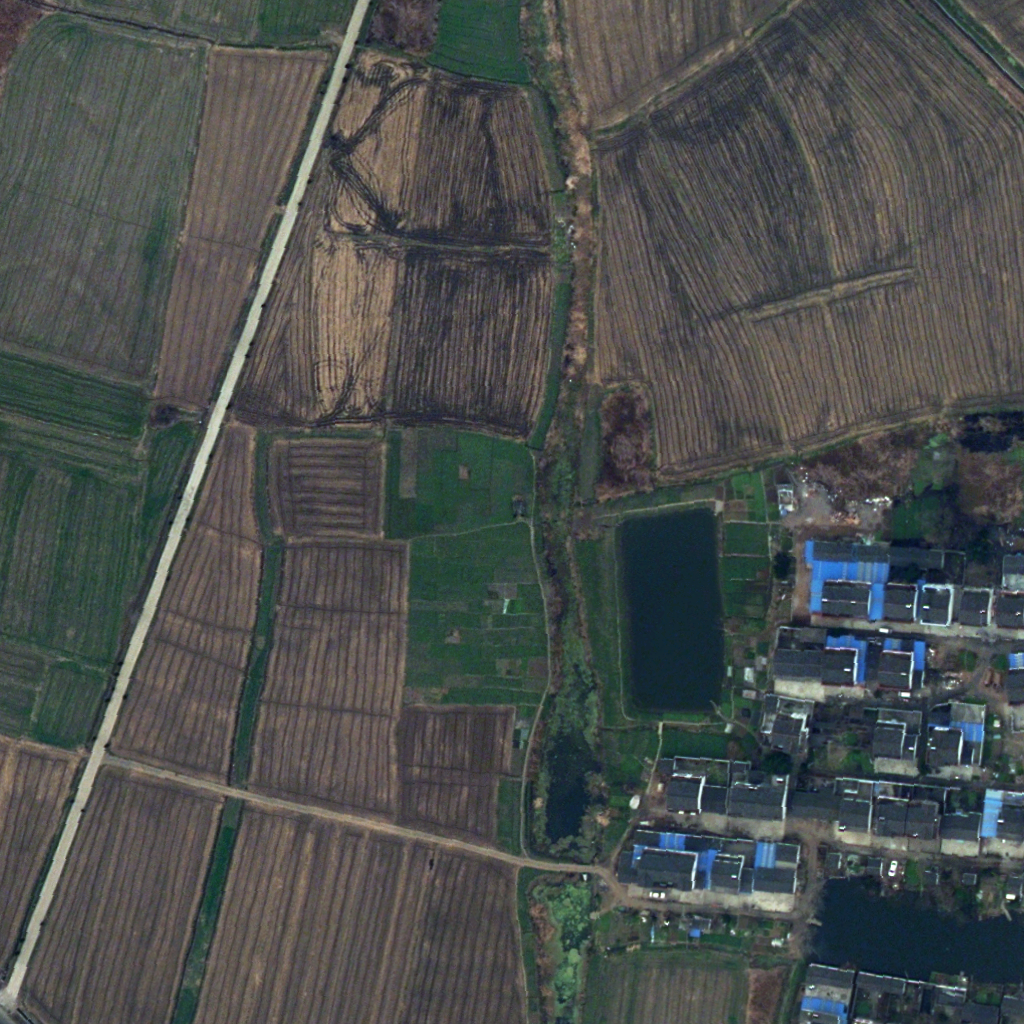}\\[4pt]
\parbox{\linewidth}{\scriptsize\raggedright\textit{What percentage of the image is covered by barren land?}\\[2pt]
Answer 3.67\%, range [1.93, 5.41]}\\[3pt]
{\scriptsize (a)}
\end{minipage}\hfill
\begin{minipage}[t]{0.31\linewidth}
\centering
\includegraphics[width=0.72\linewidth]{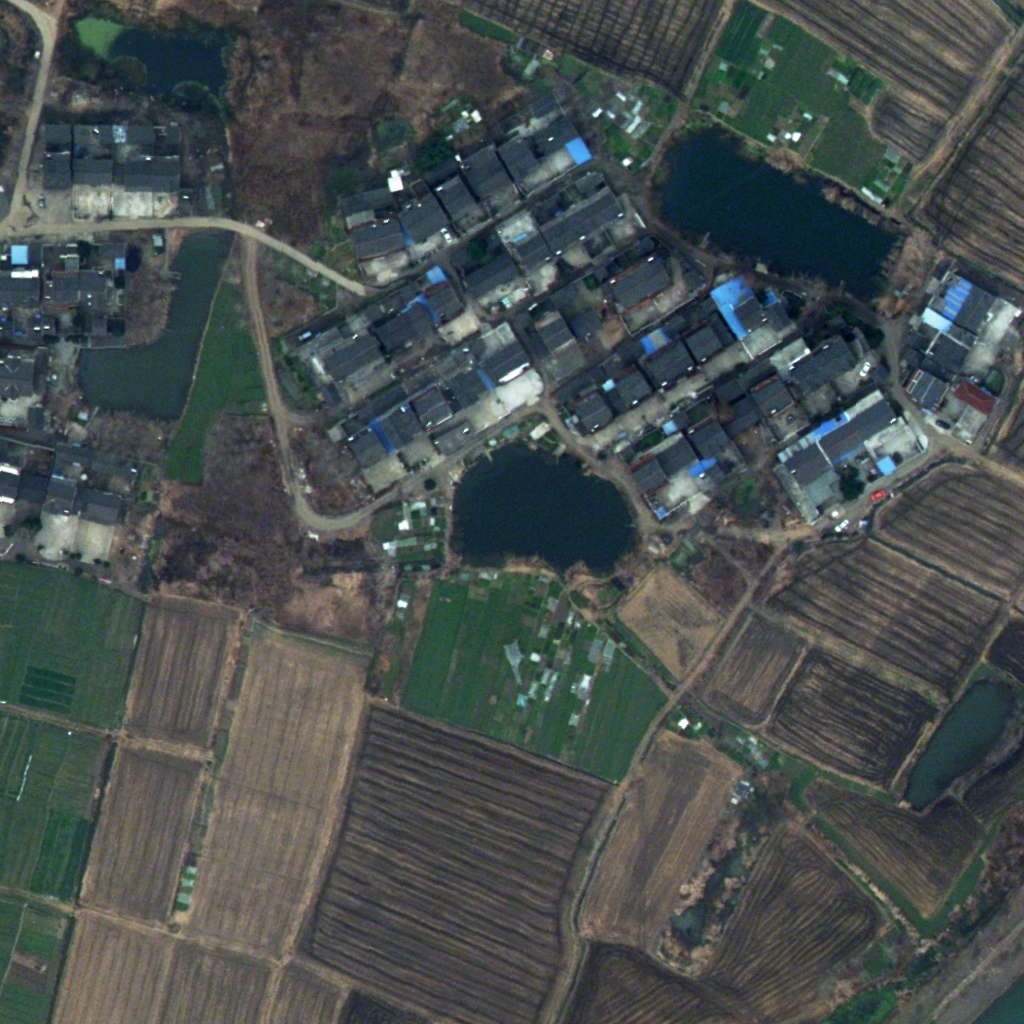}\\[4pt]
\parbox{\linewidth}{\scriptsize\raggedright\textit{How many agricultural land regions are there? When counting, ignore regions smaller than 0.125 hectares.}\\[2pt]
Answer 3, range [2, 4]}\\[3pt]
{\scriptsize (b)}
\end{minipage}\hfill
\begin{minipage}[t]{0.31\linewidth}
\centering
\includegraphics[width=0.72\linewidth]{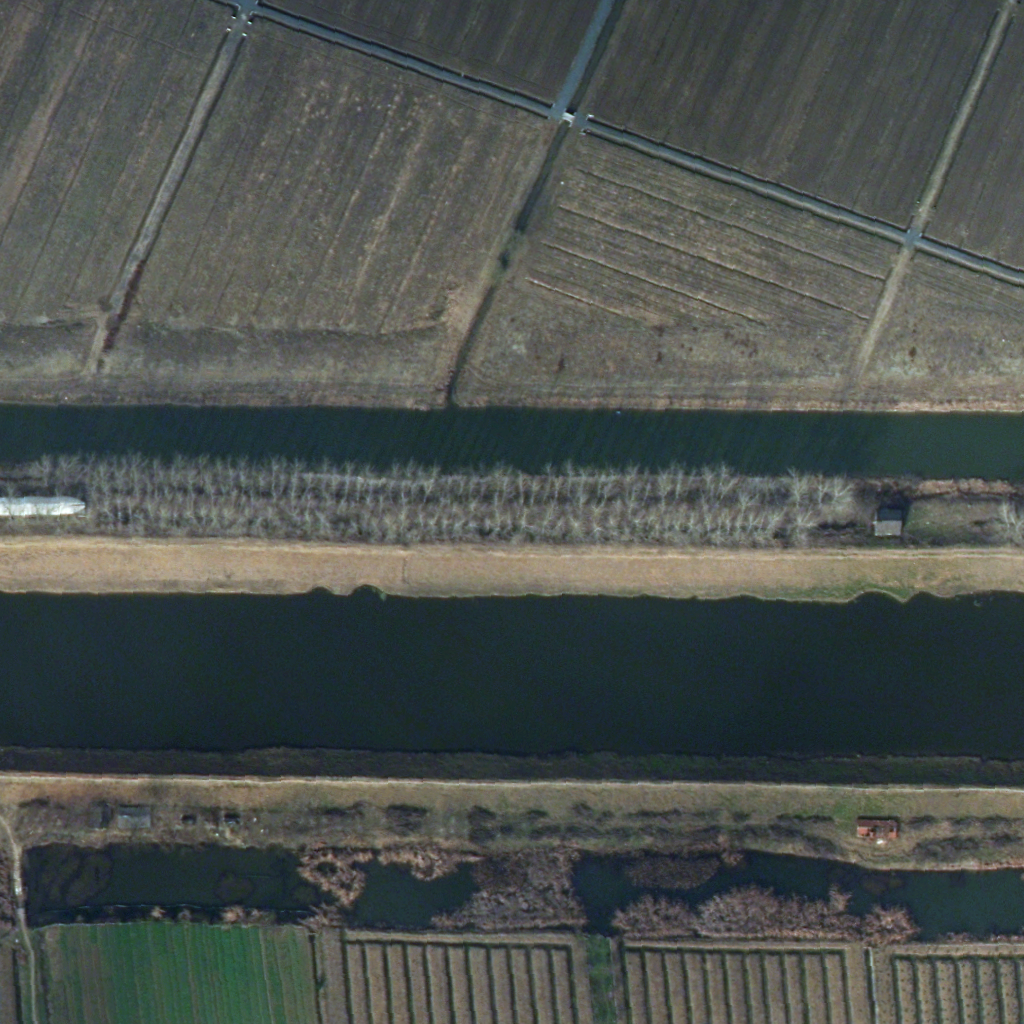}\\[4pt]
\parbox{\linewidth}{\scriptsize\raggedright\textit{Find water bodies larger than 1 hectare, then calculate how much of their area (in hectares) lies within 500\,m of barren land.}\\[2pt]
Answer 1.4\,ha, range [1.37, 1.43]}\\[3pt]
{\scriptsize (c)}
\end{minipage}
\caption{Benchmark images with example questions, reference answers and tolerance ranges.}
\label{fig:tier_visual_examples}
\end{figure}

\Cref{tab:question_examples} shows representative questions with their tolerance ranges, and \cref{fig:tier_visual_examples} shows examples on benchmark images.

\begin{table}[t]
\caption{Representative SQuID questions by tier. Tolerance ranges are informed by human annotations, and the supplementary material gives one question for each of the 24 types.}
\label{tab:question_examples}
\centering
\footnotesize
\setlength{\tabcolsep}{4pt}
\begin{tabular}{@{}>{\raggedright\arraybackslash}p{8.4cm}ccc@{}}
\toprule
\textbf{Question} & \textbf{GSD} & \textbf{Answer} & \textbf{Range} \\
\midrule
\multicolumn{4}{@{}l}{\cellcolor{blue!10}\textit{Tier 1: Basic Quantification}} \\
What percentage of the image is covered by barren land? & 0.3\,m & 33.1\% & [31.4, 34.9] \\
How many buildings are there? When counting, ignore buildings smaller than 0.01 hectares. & 0.3\,m & 4 & [3, 5] \\
Is there more water than barren land in this image? & 0.3\,m & yes & exact \\
\midrule
\multicolumn{4}{@{}l}{\cellcolor{orange!10}\textit{Tier 2: Spatial Relationships}} \\
What percentage of the urban area is within 500\,m of vegetation? & 0.5\,m & 34.35\% & [32.1, 36.6] \\
How many buildings (larger than 0.01 hectares) are within 100\,m of water (flood risk)? & 0.3\,m & 14 & [11, 17] \\
\midrule
\multicolumn{4}{@{}l}{\cellcolor{red!10}\textit{Tier 3: Complex Multi-Condition}} \\
Find barren land patches larger than 1 hectare, then calculate how much of their area (in hectares) lies within 200\,m of forest. & 0.3\,m & 1.49\,ha & [1.46, 1.52] \\
Find urban areas larger than 1 hectare, then calculate how much of their area (in hectares) lies within 100\,m of water. & 0.5\,m & 17.8\,ha & [17.4, 18.2] \\
\bottomrule
\end{tabular}
\end{table}

\subsection{Tolerance Ranges}

SQuID scores numeric answers against tolerance ranges rather than point estimates, since annotators looking at the same image count slightly different numbers and trace slightly different boundaries. We measure this variability with the median absolute deviation (MAD)~\cite{leys2013detecting} of the responses to each question,
\begin{equation}
\text{MAD} = \text{median}_i\left(|X_i - \text{median}(X)|\right),
\end{equation}
where $X = \{X_1, \ldots, X_n\}$ are the $n$ annotator responses.

For the 26 percentage questions (260 annotations), the mean MAD is $\overline{\text{MAD}}=1.735$ percentage points, and a reference percentage $P$ receives the range $[P-\overline{\text{MAD}},\,P+\overline{\text{MAD}}]$, restricted to $[0,100]$. The 9 proximity-percentage questions (90 annotations) involve distance estimation and give a larger mean MAD of 2.250 percentage points, which we use for proximity percentages.

For count questions, we normalize MAD by the median count,
\begin{equation}
\text{MAD}_c = \frac{\text{median}_i\left(|X_i - \text{median}(X)|\right)}{\text{median}(X)},
\end{equation}
which requires a positive median. The 12 count questions (120 annotations) give $\overline{\text{MAD}}_c=0.19$. A reference count $C$ receives the range $[\max(0,\lfloor C-\delta\rfloor),\,\lceil C+\delta\rceil]$ with $\delta=\max(1,\,0.19\,C)$, so small counts keep a tolerance of at least one and larger counts get wider ranges.

Answers in hectares or megawatts receive a relative tolerance of $\pm2.25\%$. This extends the numerical value of the proximity MAD as a heuristic, since annotators did not estimate these quantities directly. Categorical answers require an exact match. The calibration uses 1.0\,m NAIP imagery, and the resulting rules are applied to all sources (\cref{sec:limitations}).

\subsection{Quality Assurance}

We apply four rules. \textbf{(1) Explicit area thresholds.} Counting questions state a minimum area to reduce the effect of small segmentation artifacts. \textbf{(2) Metric consistency.} Each example includes the GSD, so that pixel measurements can be converted to meters and hectares. \textbf{(3) Zero-valued answers.} 102 questions have a zero answer, which tests whether models detect the absence of a feature. \textbf{(4) Range validation.} Reference answers and tolerance ranges must stay within physically plausible bounds.
\section{QASR (\textbf{Q}uantitative \textbf{A}gentic \textbf{S}patial \textbf{R}easoning)}
\label{sec:QASR}

QASR is the modular baseline we evaluate with predicted and reference masks (\cref{fig:eval_modes}b,c). An LLM reads the question and writes a Python program. The program calls a segmentation model to obtain masks for the requested classes, and geometric operations on these masks produce the answer. The LLM never sees the image, and areas, distances and counts are computed explicitly at the stated GSD instead of being estimated by a language model.

QASR does not preserve image information end to end. Its segmentation encoders also downsample the image, and segmentation errors can affect the measurements. What changes is where measurement happens. The masks keep their spatial indexing during the downstream computation, and the gain comes from combining these pixel-aligned outputs with explicit geometric tools. Any code-capable LLM can serve as the generator and any model that outputs pixel-level masks can serve as the segmentation model, so both can be replaced independently.

\subsection{Programs and Spatial Analysis API}
\label{subsec:code_and_api}

The LLM receives the question, the API signatures and the list of available classes. The generated program runs in a sandbox with the image path and the GSD predefined (\texttt{IMAGE\_PATH}, \texttt{gsd}) and uses an API with three functions. \texttt{segment\_\allowbreak image\_\allowbreak from\_\allowbreak path} returns the masks of the requested classes as shapes with pixel and hectare areas, \texttt{find\_\allowbreak shapes\_\allowbreak within\_\allowbreak distance} clips target shapes to their parts within a given distance of reference shapes, and \texttt{calculate\_\allowbreak shape\_\allowbreak distances} computes minimum distances between shapes. For ``Find water bodies larger than 1 hectare, then calculate how much of their area (in hectares) lies within 200\,m of agricultural land'', a program is

\begin{lstlisting}[language=Python, basicstyle=\scriptsize\ttfamily, frame=single]
result = segment_image_from_path(IMAGE_PATH, ["water", "agric"], gsd=gsd)
water = [s for s in result["shapes"] if s["class_type"] == "water"]
agric = [s for s in result["shapes"] if s["class_type"] == "agric"]
large_water = [s for s in water if s["area_hectares"] > 1.0]
within = find_shapes_within_distance(large_water, agric, 200.0, gsd)
answer = sum(s["area_hectares"] for s in within)
\end{lstlisting}
The supplementary material gives the full API, the developer prompt and further examples.

\textbf{Role of the LLM.} The segmentation model handles perception. The LLM maps the question to a program. It selects the classes and thresholds, chooses the geometric operations, orders them and writes the final aggregation. The API then performs the deterministic computation on the masks. GPT-5 is not required. It is the strongest generator we tested, and generator quality still matters when using reference masks (\cref{sec:experiments}). Because SQuID questions come from templates, a hand-written question-to-program parser would also work for the current benchmark. We use an LLM to avoid writing such a parser.

\subsection{Segmentation Models}
\label{subsec:seg_models}

\begin{figure}[t]
\centering
\setlength{\tabcolsep}{2pt}
\begin{tabular}{@{}cc@{}}
\includegraphics[width=0.49\linewidth,trim=0 0 0 500pt, clip]{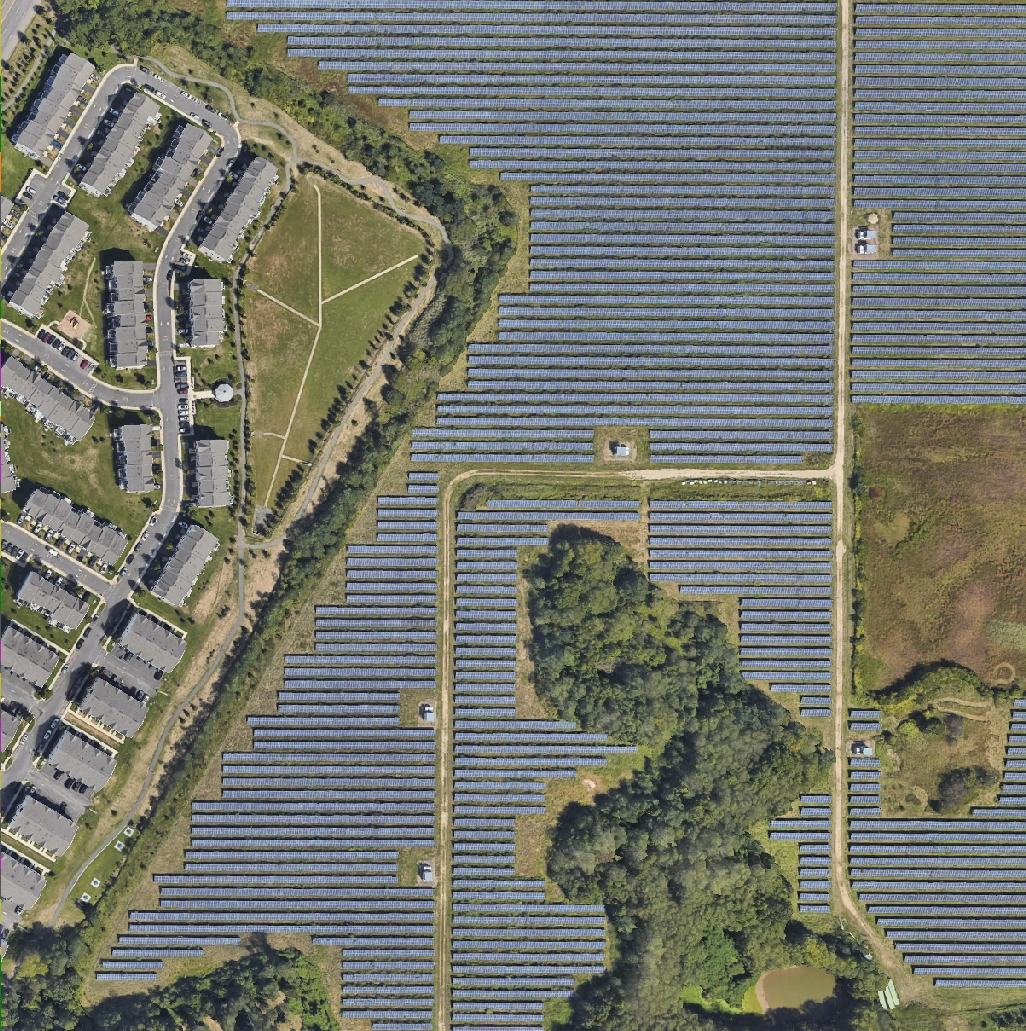} &
\includegraphics[width=0.49\linewidth,trim=0 0 0 500pt, clip]{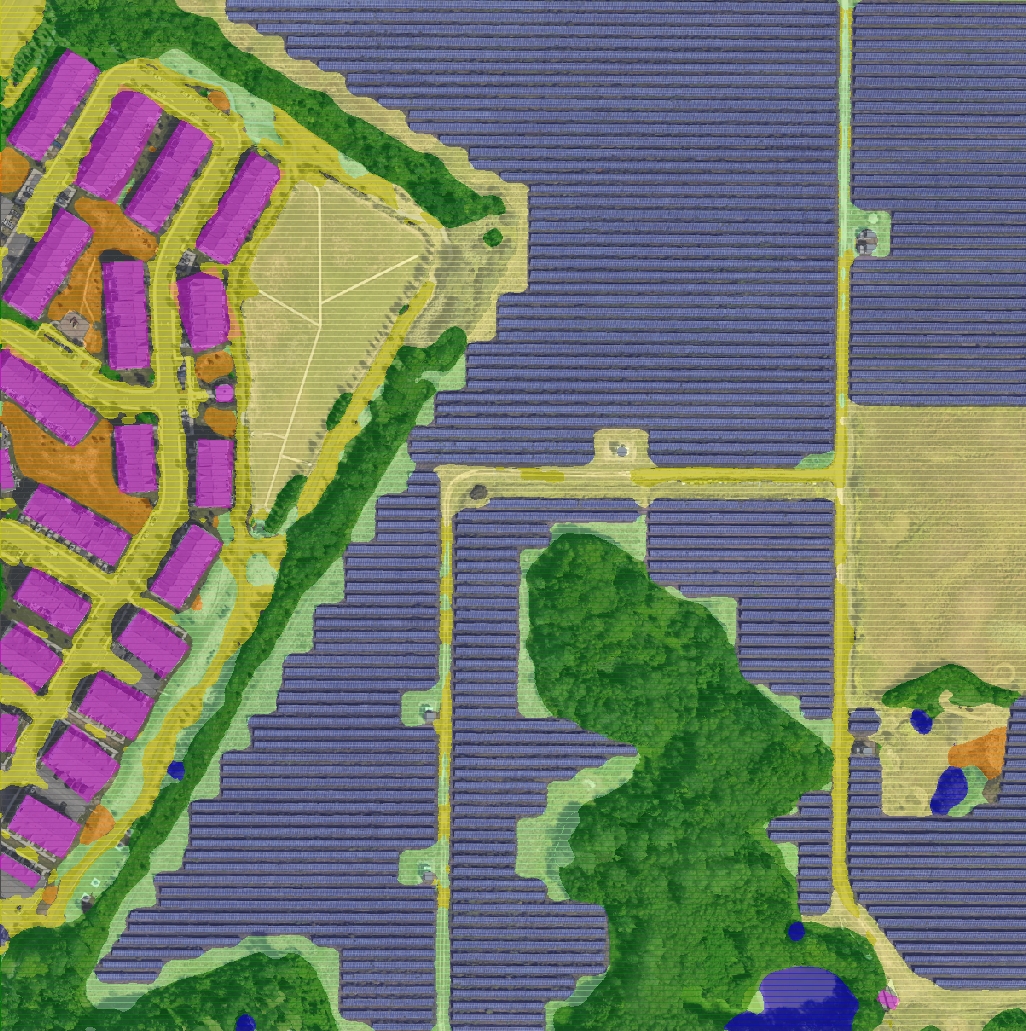} \\
\end{tabular}
\caption{\textbf{ConvNeXt-UNet segmentation.} Input image (left) and combined output (right), with forest in dark green, agriculture in yellow, grass in light green, barren land in orange, water in blue, solar panels in dark blue and buildings in magenta.}
\label{fig:seg_example}
\end{figure}

\textbf{ConvNeXt-UNet.} We train separate ConvNeXt~\cite{liu2022convnet} encoder and U-Net~\cite{ronneberger2015unetconvolutionalnetworksbiomedical} decoder models on several public RS segmentation datasets. Each model is trained at several resolutions, and inputs outside its GSD range are rescaled into it before inference. At inference, each pixel receives the class with the largest logit across models, with per-class weights for overlapping classes and a mode filter to reduce noise (\cref{fig:seg_example}). New classes can be added by adding models. Training data and per-model scores are in the supplementary material.

\textbf{DINOv3-Mask2Former.} To test a different backbone, we pair a frozen DINOv3 ViT-L/16~\cite{simeoni2025dinov3} pretrained on SAT-493M satellite imagery with a Mask2Former decoder~\cite{cheng2022mask2former}. An adapter following~\cite{chen2023vitadapter} combines features from layers 4, 11, 17 and 23 into maps at strides 4, 8, 16 and 32 for a Mask2Former head with 100 queries and hidden dimension 256. Trained on EarthVQA, it reaches 57.6\% validation mIoU across 8 classes, including background. Training details are in the supplementary material.

For both models, the EarthVQA checkpoint was selected on the EarthVQA validation split, which contains the EarthVQA benchmark images, so their EarthVQA results may be slightly optimistic.
\section{Experimental Results}
\label{sec:experiments}

\subsection{Setup}

\textbf{Models.} QASR runs with four segmentation models, ConvNeXt-UNet, DINOv3-Mask2Former, RemoteSAM~\cite{yao2025remotesam} and SAM-3~\cite{carion2025sam3}, and three program generators, GPT-5~\cite{openai2025gpt5}, gpt-oss-120B~\cite{openai2025gptoss120bgptoss20bmodel} and Llama-3.1-8B~\cite{meta2024llama3}. The end-to-end VLMs are GPT-5, Qwen3-VL-30B-A3B-Thinking~\cite{qwen3technicalreport}, Qwen3-VL-8B, Qwen2.5-VL-7B~\cite{bai2025qwen25vl}, CoVT~\cite{qin2025chain} and three RS-specialized VLMs, GeoChat~\cite{kuckreja2023geochatgroundedlargevisionlanguage}, Geo-R1~\cite{georl2025} and EarthDial-4B~\cite{soni2025earthdial}. We also fine-tune Qwen2.5-VL-7B-Instruct on SQuID-Train with LoRA~\cite{hu2022lora} (\cref{sec:squid_train}).

\textbf{Metrics.} A numeric answer is correct when it falls within its tolerance range, and a categorical answer when it matches the reference exactly. We report accuracy per tier and overall. Questions can share an image, so differences of one or two points should be read with care.

\textbf{Protocol.} Except for the fine-tuned model, question answering and program generation are evaluated zero-shot, with pass@1 and greedy decoding (temperature 0) where supported. GPT-5 uses minimal reasoning effort with no explicit output-token cap. In end-to-end mode, it receives high-detail images. gpt-oss-120B uses medium effort with a budget of 4,096 tokens. Qwen3-VL-30B-A3B-Thinking has a reasoning budget of 4,096 tokens, after which a closing thinking token is appended and 10 more tokens are allowed for the answer. The QASR developer prompt is in the supplementary material.

\begin{table*}[t]
\centering
\caption{Accuracy (\%) on SQuID by tier. Among the rows evaluated on the full benchmark, the best result is in \textbf{bold} and the second best is \underline{underlined}.}
\label{tab:results}
\scriptsize
\resizebox{\textwidth}{!}{
\begin{tabular}{clccrrrc}
\toprule
\multicolumn{2}{c}{\multirow{2}{*}{Method}} & \multirow{2}{*}{Base LLM} & \multirow{2}{*}{Segmentation} & \multicolumn{3}{c}{$\uparrow$ Accuracy (\%)} & \multirow{2}{*}{$\uparrow$ Overall (\%)} \\
\cmidrule(lr){5-7}
\multicolumn{2}{c}{} & & & Tier 1 & Tier 2 & Tier 3 & \\
\midrule
\multirow{6}{*}{\begin{tabular}{c}Modular,\\predicted masks\end{tabular}}
& QASR-A & GPT-5        & ConvNeXt  & \best{53.52} & \underline{54.06} & \best{18.84} & \best{42.00} \\
& QASR-B & gpt-oss-120B & ConvNeXt  & 44.79 & 45.94 &  6.08 & 32.10 \\
& QASR-C & GPT-5        & DINOv3    & \underline{50.99} & 47.40 & 12.46 & 36.90 \\
& QASR-D & Llama-3.1-8B & ConvNeXt  & 39.86 & 41.88 &  5.79 & 29.00 \\
& QASR-E & GPT-5        & RemoteSAM & 23.52 & 38.64 &  6.08 & 22.30 \\
& QASR-F & GPT-5        & SAM-3     & 27.61 & 31.98 & 11.28 & 23.45 \\
\midrule
\multirow{8}{*}{\begin{tabular}{c}End-to-end,\\zero-shot\end{tabular}}
& GPT-5      & GPT-5            & - & 39.30 & 34.09 & 10.83 & 28.10 \\
& Qwen2.5-VL & Qwen2.5-7B       & - & 22.82 & 25.65 &  0.30 & 16.10 \\
& Qwen3-VL   & Qwen3-8B         & - & 28.59 & 36.85 & 10.09 & 24.90 \\
& Qwen3-VL   & Qwen3-30B-A3B    & - & 39.01 & 36.85 &  3.71 & 26.45 \\
& Geo-R1     & Qwen2.5-3B       & - & 31.41 & 28.57 &  1.48 & 20.45 \\
& GeoChat    & Vicuna-1.5-7B    & - & 35.92 & 37.01 &  0.30 & 24.25 \\
& EarthDial  & Phi-3-mini (4B)  & - & 32.82 & 50.16 &  0.89 & 27.40 \\
& CoVT       & Qwen2.5-7B       & - & 33.38 & 25.32 &  0.74 & 19.90 \\
\midrule
\begin{tabular}{c}End-to-end,\\SQuID-Train\end{tabular}
& Qwen2.5-VL (LoRA) & Qwen2.5-7B & - & \best{53.52} & \best{56.01} & \underline{13.20} & \underline{40.70} \\
\midrule
\multirow{2}{*}{\begin{tabular}{c}Modular,\\reference masks$^\dagger$\end{tabular}}
& QASR-A$^\dagger$ & GPT-5        & Reference masks & 93.2 & 93.7 & 62.0 & 82.6 \\
& QASR-D$^\dagger$ & Llama-3.1-8B & Reference masks & 83.3 & 81.3 & 38.3 & 67.1 \\
\bottomrule
\end{tabular}
}
\vspace{2pt}
{\scriptsize Tier~1 is basic quantification, Tier~2 spatial relationships and Tier~3 complex multi-condition. $^\dagger$Reference-mask mode on the 1,950 mask-derived questions, where the same programs score 42.2\% (GPT-5) and 29.1\% (Llama-3.1-8B) with predicted masks. These rows are not ranked.\par}
\end{table*}

\subsection{Main Results}

\Cref{tab:results} reports all systems. QASR with GPT-5 and ConvNeXt reaches 42.0\%, against 28.1\% for GPT-5 used end to end, a gain of 13.9 percentage points. It improves every tier, by 14.2 points on Tier~1, 20.0 on Tier~2 and 8.0 on Tier~3. The largest gains are on question types that reduce to explicit geometry on masks, such as fragmentation (81.63\% vs 26.53\%), connectivity (74.04\% vs 37.50\%), proximity percentages (40.65\% vs 19.51\%) and counts (56.74\% vs 36.52\%).

\textbf{Where the gain comes from.} Open-vocabulary masks alone do not guarantee a gain. QASR with RemoteSAM (22.30\%) or SAM-3 (23.45\%) scores below GPT-5 used end to end (28.10\%). The gain depends on the segmentation and program-generation configuration. It is a result for the complete system, not evidence of better reasoning inside the language model.

\textbf{Remote sensing VLMs.} RS specialization does not close the gap. EarthDial-4B, trained on 11.11M RS instruction pairs, reaches 27.4\% (32.8\%, 50.2\% and 0.9\% on Tiers~1 to~3), close to GPT-5. It has the highest Tier~2 score among the evaluated zero-shot end-to-end VLMs, including 91.8\% on fragmentation, yet GeoChat, Geo-R1 and EarthDial all stay below 2\% on Tier~3.

\textbf{Training on SQuID-Train.} Fine-tuning Qwen2.5-VL-7B on SQuID-Train raises its accuracy from 16.1\% to 40.7\%, close to QASR overall and on Tiers~1 and~2 (53.5\% and 56.0\%), but only 13.2\% on Tier~3. Training on question-answer pairs therefore helps end-to-end models substantially without solving multi-condition questions. SQuID-Train shares templates, sources and GSDs with the benchmark, so this result measures in-distribution learning, and part of the gain may come from answer priors of the templates (\cref{sec:limitations}).

\textbf{Human reference.} We score each of the 500 annotator responses on the 50 NAIP questions with the same rule as a model output (\cref{sec:human_ref}). Of these, 313 are accepted, so individual annotators agree with the consensus on 62.6\% of their responses, while the 13 systems scored on these questions reach 12 to 46\%. The consensus defines the reference for these questions, so 62.6\% measures agreement rather than an independent upper bound, and the annotators were trained researchers rather than RS experts.

\subsection{Perception versus Program Errors}
\label{subsec:replay}

\begin{figure}[t]
\centering
\setlength{\tabcolsep}{2pt}
\begin{tabular}{cc}
    \includegraphics[width=0.48\linewidth]{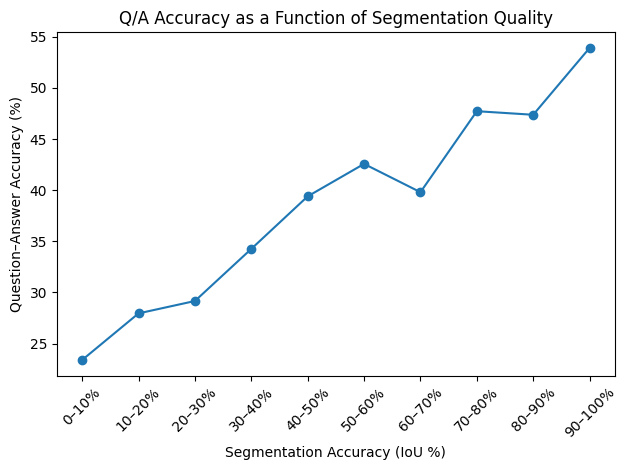} &
    \includegraphics[width=0.48\linewidth]{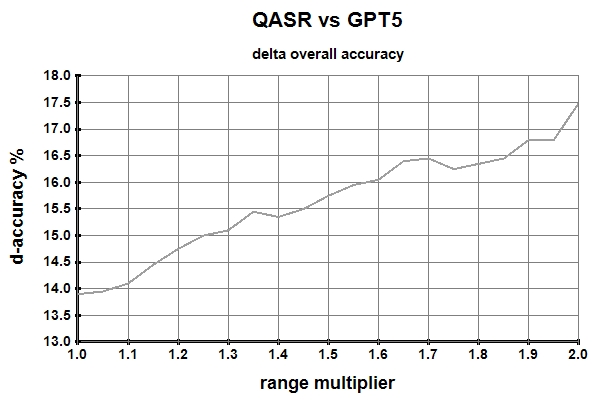}
\end{tabular}
\caption{\textbf{Left.} Accuracy of QASR (GPT-5 + ConvNeXt) grouped by the per-image IoU between predicted and reference masks. QASR uses predicted masks in this analysis. \textbf{Right.} Accuracy advantage of QASR over GPT-5 used end to end, in percentage points, as numeric tolerance ranges are widened by factors from 1.0 to 2.0. The gap grows from 13.9 to 17.5 points.}
\label{fig:results2}
\end{figure}

Only 2 of the 2,000 QASR programs (0.1\%) fail with a syntax error, with no retries. To measure the effect of segmentation, we use the reference-mask mode (\cref{subsec:companion}). For each of the 1,950 mask-derived questions, we rerun the original GPT-5 program with reference masks, without a new LLM call or changes to the API or scoring rule (\cref{sec:gt_replay}). Accuracy rises from 42.2\% to 82.6\%, with 93.2\%, 93.7\% and 62.0\% on Tiers~1 to~3 (\cref{tab:results}). Of the 1,127 questions answered incorrectly with predicted masks, 814 become correct and 313 stay incorrect, while 27 correct answers become incorrect (\cref{tab:oracle}). Segmentation is therefore the main bottleneck in this configuration, although replacing masks does not improve every answer. With the same reference masks, Llama-3.1-8B programs reach 67.1\% overall, 15.5 points below GPT-5, and 38.3\% on Tier~3. The program generator therefore still matters when the masks are held fixed. Accuracy also tends to increase with per-image segmentation IoU (\cref{fig:results2}, left).

\begin{table}[t]
\centering
\caption{Outcome of each of the 1,950 mask-derived questions when the same QASR programs run on predicted and on reference masks.}
\label{tab:oracle}
\small
\begin{tabular}{llrr}
\toprule
Predicted masks & Reference masks & GPT-5 & Llama-3.1-8B \\
\midrule
Correct & Correct & 796 & 533 \\
Correct & Incorrect & 27 & 34 \\
Incorrect & Correct & 814 & 776 \\
Incorrect & Incorrect & 313 & 607 \\
\bottomrule
\end{tabular}
\end{table}

\subsection{Ablations}

\textbf{Program generator.} With ConvNeXt masks, gpt-oss-120B reaches 32.1\% and Llama-3.1-8B 29.0\%, against 42.0\% for GPT-5. With Llama-3.1-8B, QASR is close to GPT-5 used end to end (29.0\% vs 28.1\%).

\textbf{Segmentation model.} With GPT-5 as program generator, replacing ConvNeXt by DINOv3 lowers accuracy from 42.0\% to 36.9\%, and RemoteSAM and SAM-3 reach 22.30\% and 23.45\%. These configurations differ in training data and class definitions as well as architecture. The source-specific ConvNeXt models are trained with the class definitions of the questions (81.2\% and 80.0\% F1 on EarthVQA and DeepGlobe, 97.0\% F1 on solar panels and 98.4\% Dice on AIRS roofs). DINOv3 keeps its SAT-493M backbone frozen and trains only an adapter and decoder on EarthVQA (57.6\% mIoU). RemoteSAM is trained on RemoteSAM-270K, whose 297 categories follow a different taxonomy, and SAM-3 is used zero-shot, so their masks can look plausible and still disagree with the class boundaries of the source datasets. Cross-domain results on GEOBench-VLM are in the supplementary material (\cref{sec:crossdomain}).

\textbf{Tolerance ranges.} Widening numeric tolerance ranges by a factor of 2 raises QASR from 42.0\% to 51.4\% and GPT-5 from 28.1\% to 33.9\% (\cref{fig:results2}, right). The gap grows from 13.9 to 17.5 percentage points, and QASR remains ahead over the tested range.
\section{Limitations}
\label{sec:limitations}

\textbf{Coverage.} SQuID is geographically concentrated in East, Southeast and South Asia and the United States, and models trained in one region often degrade in others~\cite{kalluri2023geonet}. It uses a closed class vocabulary and covers GSDs from 0.3 to 1.0\,m only, while what can be measured depends on resolution~\cite{woodcock1987factor}. We have not tested scale shifts outside this range, and QASR handles inputs outside the training range of its segmentation models by rescaling. SQuID only asks questions whose answers can be derived from masks, so elevation, temporal change and non-geometric instance attributes remain outside its scope.

\textbf{Question form.} Questions are template-generated. A parser implementing the answer-generation rules could solve these questions using the reference masks, so that mode is informative for general-purpose program generators rather than template-specific systems. Templates can also make answer priors learnable~\cite{goyal2017making}, which may account for part of the gain from SQuID-Train. Some multi-condition templates admit more than one reading, for example whether a region is clipped to the buffer before its area is summed, so part of the remaining Tier~3 error with reference masks may reflect ambiguity. Binary answers are not yet balanced within each question type, Tier~3 is dominated by one template family (490 of 674 questions), and several types contain fewer than 20 questions.

\textbf{Reference answers and tolerances.} Mask-derived answers inherit errors in the source masks, and label errors in test sets can change model rankings~\cite{northcutt2021pervasive}. Tolerance calibration uses 47 numeric questions on 1.0\,m NAIP imagery, and its rules are applied to 0.3 and 0.5\,m imagery and extended heuristically to hectare and megawatt answers. Proximity questions require more visual estimation and have lower agreement than count questions ($\alpha$ of 0.42 versus 0.96). Our annotators were trained researchers, not an independent RS-expert cohort, and expertise can affect land-cover identification~\cite{see2013comparing}.

\textbf{QASR.} QASR depends on segmentation quality and a fixed geometry API. In our evaluation, it performs best with in-domain closed-vocabulary segmentation models and is slower than an end-to-end call (7.6\,s vs 5.2\,s per sample), with higher total token usage (1,650 vs 682 per sample).

Future work includes more regions and GSDs, more classes, paraphrased questions, better-balanced binary answers, a larger Tier~3 and a larger human-validation set with RS experts.

\section{Conclusion}
\label{sec:conclusion}

We presented SQuID, a benchmark of 2,000 questions over 587 satellite images for quantitative spatial reasoning, with tolerance ranges informed by human annotations, reference masks and a training split on disjoint images (SQuID-Train). The evaluated zero-shot end-to-end VLMs, including RS-specialized models, reach at most 28.1\%. QASR, a modular baseline that measures on pixel-aligned masks, reaches 42.0\%, and running its programs on the reference masks raises accuracy from 42.2\% to 82.6\% on the 1,950 mask-derived questions, so segmentation is the main bottleneck in this configuration. Fine-tuning an end-to-end model on SQuID-Train reaches 40.7\%. Multi-condition questions remain the hardest in every setting, at 62.0\% for QASR with reference masks and 13.2\% after fine-tuning. We hope SQuID will support end-to-end models that learn metric spatial reasoning, and that its evaluation modes will help assess the effects of segmentation and program generation.


\bibliographystyle{plainnat}
\bibliography{main}


\newpage
\appendix
\newpage
\raggedbottom
\makeatletter
\setlength{\@fptop}{0pt}
\setlength{\@fpsep}{20pt plus 4pt}
\setlength{\@fpbot}{0pt plus 1fil}
\makeatother
\renewcommand{\floatpagefraction}{0.5}

\section{Supplementary Material}
\label{sec:supp}

This appendix gives the complete versions of the tables and figures that are shortened in the main paper, the training details of the segmentation models, a cross-domain evaluation, and the protocols behind the reference-mask replay, SQuID-Train and the human reference. The text comes first, and the tables and figures follow in this order.

\begin{itemize}[itemsep=1pt, topsep=3pt, leftmargin=1.5em]
\item \Cref{tab:convnext_training}: ConvNeXt-UNet training configuration.
\item \Cref{tab:QASR_classes}: topics QASR can request and the ConvNeXt models behind them.
\item \Cref{tab:dinov3_results}: DINOv3 + Mask2Former configuration and results.
\item \Cref{tab:geobench}: cross-domain evaluation on GEOBench-VLM.
\item \Cref{fig:annotation_interface}: annotation interface used for the human calibration.
\item \Cref{tab:classes_supp}: classes of each source and their minimum area thresholds.
\item \Cref{tab:questions_supp}: one complete question for each of the 24 types.
\item \Cref{tab:types_supp}: the 24 question types and their counts.
\item \Cref{fig:developer_prompt}: QASR developer prompt.
\item \Cref{fig:code_examples_supp}: programs for a percentage, a counting and a proximity question.
\item \Cref{fig:full_pipeline}: a full QASR pipeline example.
\item \Cref{fig:gpt_oss_examples}: a correct and an incorrect prediction from gpt-oss-120B programs.
\item \Cref{tab:full_results_by_type}: accuracy by question type for 12 configurations.
\item \Cref{fig:qwen_thinking_traces}: reasoning traces of Qwen3-VL-30B-A3B-Thinking.
\end{itemize}

\subsection{Segmentation Models}
\label{sec:convnext_details}

\textbf{ConvNeXt-UNet.} The main segmentation model of QASR uses a ConvNeXt~\cite{liu2022convnet} encoder pretrained on ImageNet with a U-Net~\cite{ronneberger2015unetconvolutionalnetworksbiomedical} decoder. We train separate models on EarthVQA, DeepGlobe, Solar and AIRS~\cite{chen2018aerial} (\cref{tab:convnext_training}). Training examples are generated by random affine patching at several resolutions spanning a range of GSDs, followed by random color augmentation (RGB shifts, brightness, contrast and saturation changes, gamma correction). An input image outside the trained GSD range is rescaled to the center of the range before inference. For instance segmentation of building roofs (AIRS), we use the same architecture with a background class for non-object pixels.

At inference, the segmentation server receives an image and a list of requested topics, which are semantic classes or object instances. Each pixel receives the class with the largest logit across models, per-class weights reflect confidence or disable classes of individual models, and a mode box filter assigns the most common class within a local neighborhood to reduce noise. The list of available topics is given to the program generator (\cref{tab:QASR_classes}).

\textbf{DINOv3 + Mask2Former.}\label{sec:dinov3_details} As an alternative, we train a DINOv3 + Mask2Former model following the official DINOv3 protocol~\cite{simeoni2025dinov3}, with a frozen satellite-pretrained ViT-L/16 backbone (SAT-493M, about 300M parameters), a trainable adapter and a Mask2Former decoder (56.6M parameters). It reaches 57.58\% mIoU on the EarthVQA validation split (\cref{tab:dinov3_results}). Its lower accuracy on SQuID aside, it shows that QASR accepts a different segmentation model without modification.

\subsection{Cross-Domain Evaluation}
\label{sec:crossdomain}

To evaluate QASR beyond SQuID, we use the \texttt{Single} split of GEOBench-VLM~\cite{danish2025geobenchvlmbenchmarkingvisionlanguagemodels}, with 3,211 multiple-choice questions across 17 task types. For QASR, counting tasks (1,054 questions) use program generation with segmentation, while the other tasks (2,157 questions) use GPT-5 end to end. GPT-5 and the two QASR configurations therefore share the same accuracy on the non-counting tasks (57.3\%).

\Cref{tab:geobench} reports the results. GPT-5 used end to end reaches 48.2\% overall. On counting, QASR with SAM-3 (15.8\%) and RemoteSAM (15.6\%) both fall below GPT-5 (29.6\%). Performance varies by object category. On vehicle counting, QASR with RemoteSAM reaches 50.7\% against 18.0\% for GPT-5, but it scores lower on buildings and aircraft. These results do not isolate the effects of object size, training data or class definitions.

\subsection{Human Annotation and Dataset Details}
\label{sec:dataset_details}

\begin{figure}[tp]
\centering
\begin{subfigure}[t]{0.44\linewidth}
    \centering
    \includegraphics[width=\textwidth]{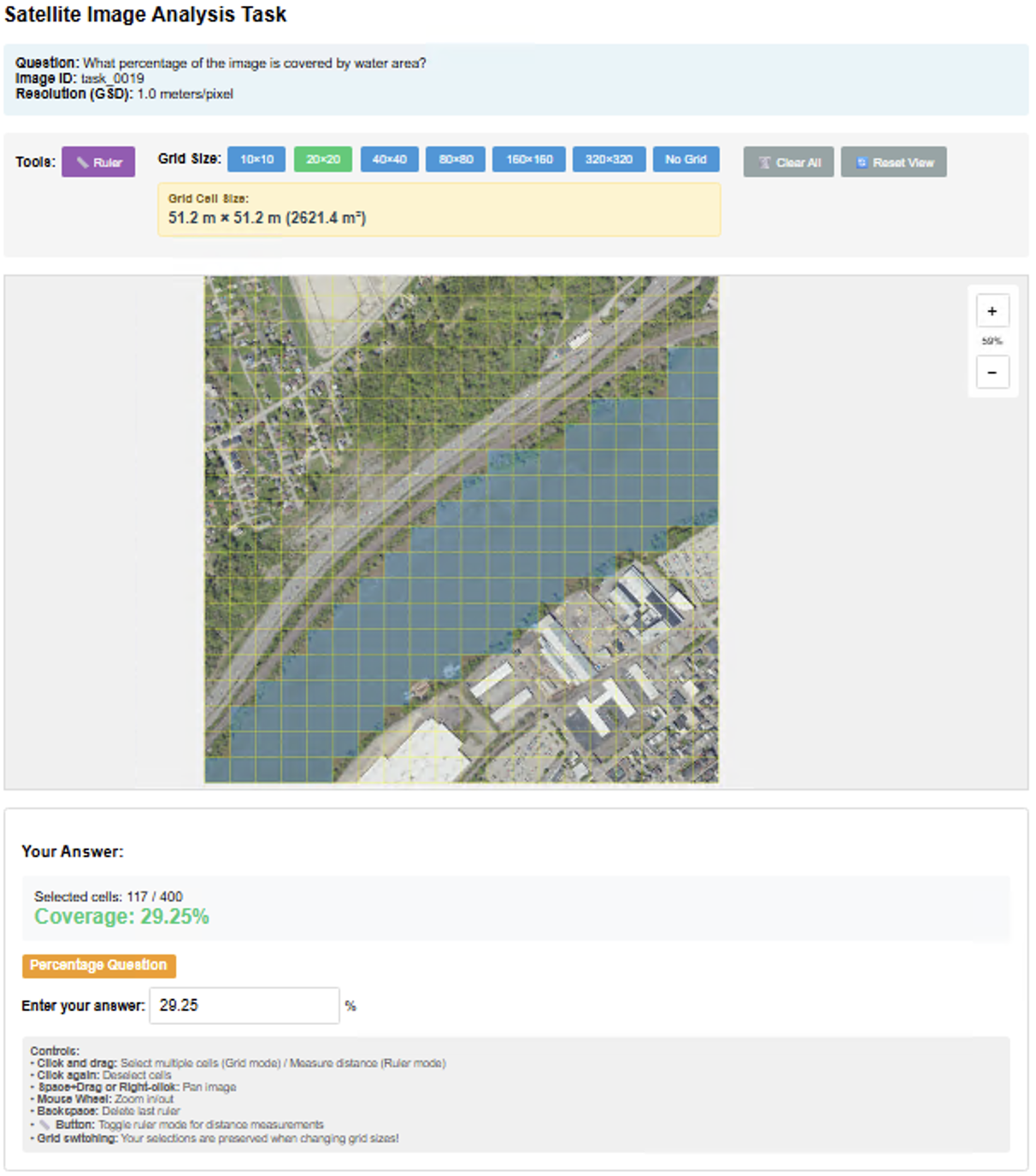}
    \caption{Grid cell selection}
    \label{fig:annotation_grid}
\end{subfigure}
\hfill
\begin{subfigure}[t]{0.44\linewidth}
    \centering
    \includegraphics[width=\textwidth]{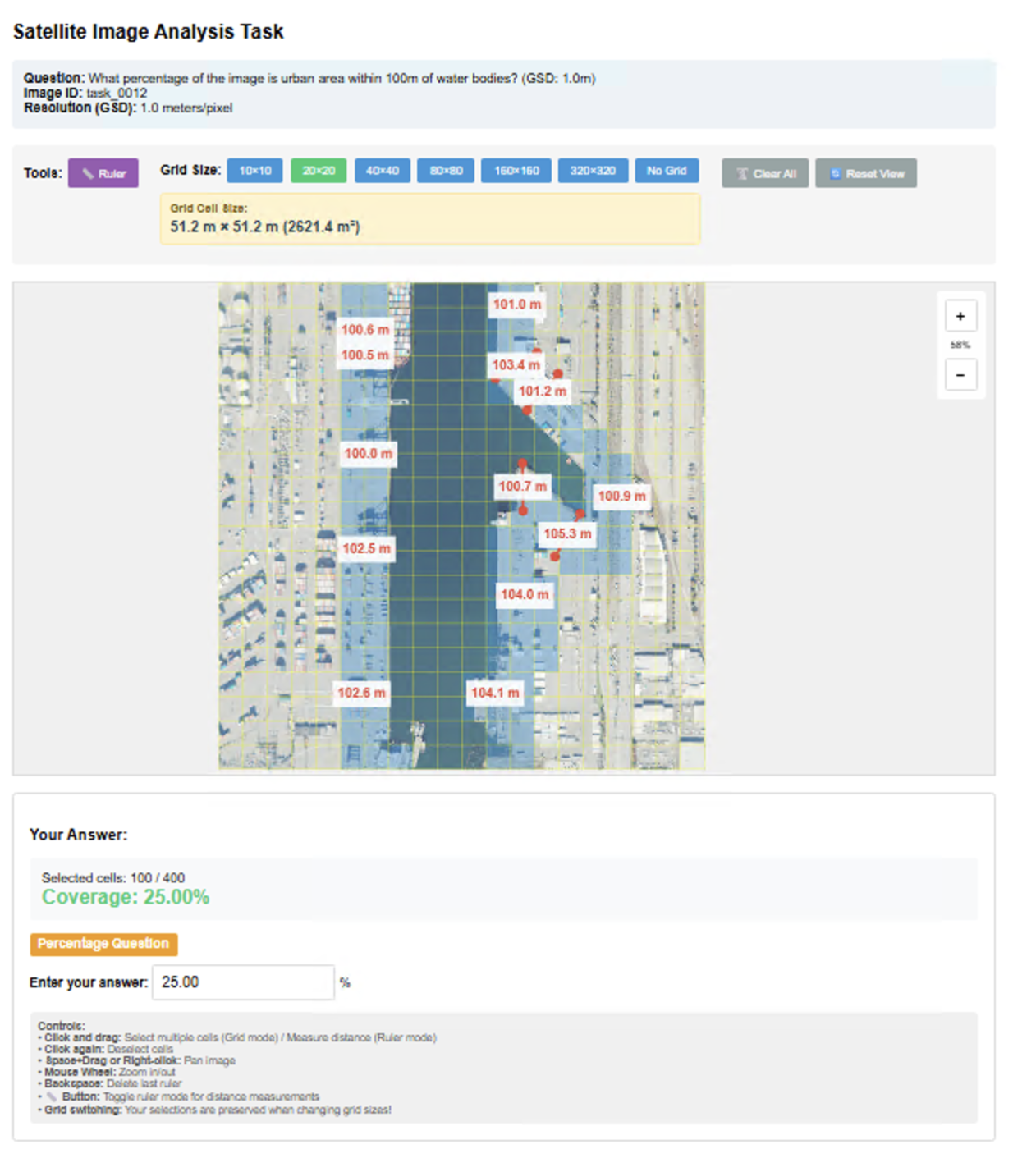}
    \caption{Distance ruler}
    \label{fig:annotation_ruler}
\end{subfigure}
\caption{\textbf{Annotation interface} built on Turkle~\cite{turkle2023}. \textbf{(a)} Annotators select the grid cells that cover a land-cover class to answer percentage questions, with a grid adjustable from 10$\times$10 to 320$\times$320 cells. \textbf{(b)} A distance ruler supports proximity questions. Ten annotators used this interface to produce 500 annotations for 50 questions over 14 NAIP images at 1.0\,m GSD.}
\label{fig:annotation_interface}
\end{figure}

Ten annotators answered the 50 NAIP questions with a grid-based interface built on Turkle~\cite{turkle2023} (\cref{fig:annotation_interface}). For percentage questions, they select the grid cells that cover a land-cover class, with a grid adjustable from 10$\times$10 to 320$\times$320 cells, and for proximity questions they measure distances with a ruler. \Cref{tab:classes_supp} lists the classes of each source with the minimum area stated in the questions, \cref{tab:questions_supp} gives one complete question per type, and \cref{tab:types_supp} gives the number of questions of each type.

\subsection{QASR Prompt and Programs}
\label{sec:qasr_prompt}

\begin{figure}[tp]
\centering
\begin{minipage}[t]{0.38\linewidth}
\vspace{0pt}
\centering
{\footnotesize SQuID image}\\[2pt]
\includegraphics[width=\linewidth]{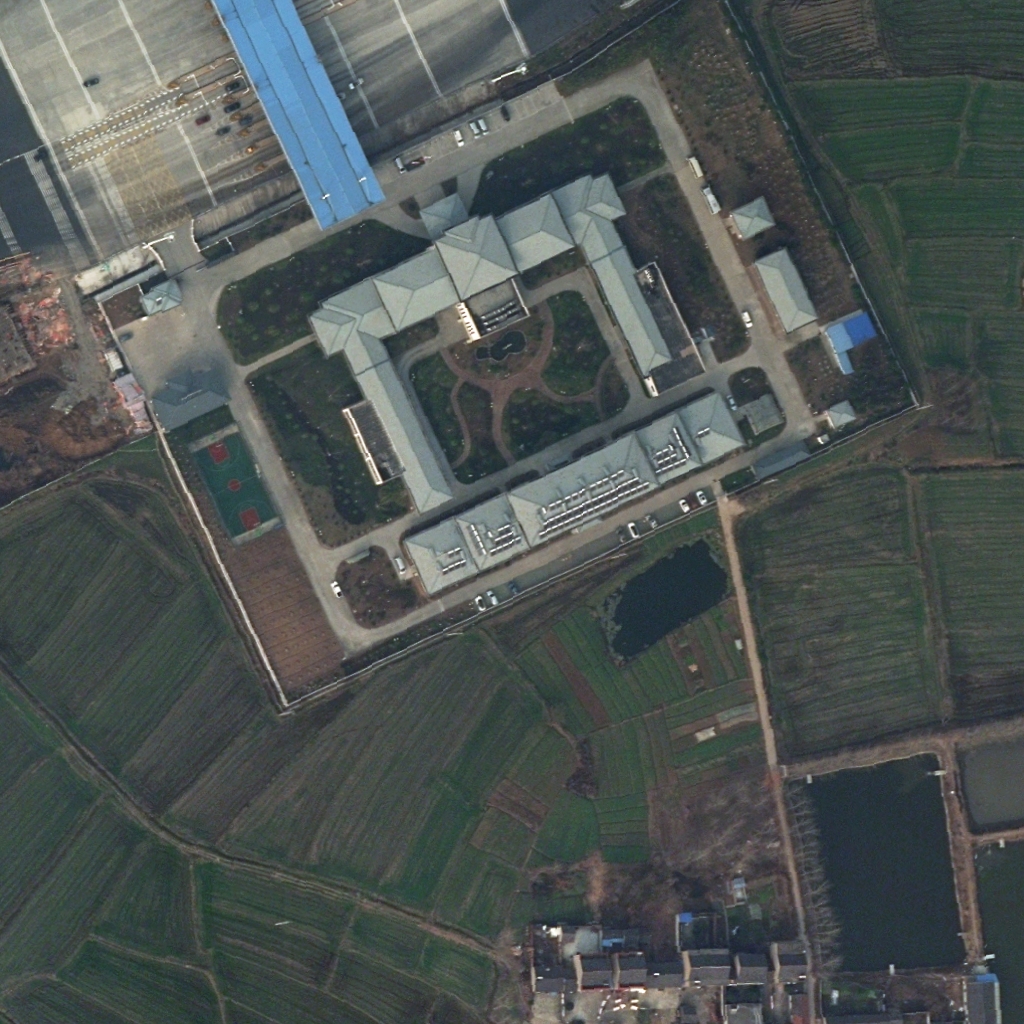}\\[6pt]
{\footnotesize Predicted agricultural land (yellow) and roofs (magenta)}\\[2pt]
\includegraphics[width=\linewidth]{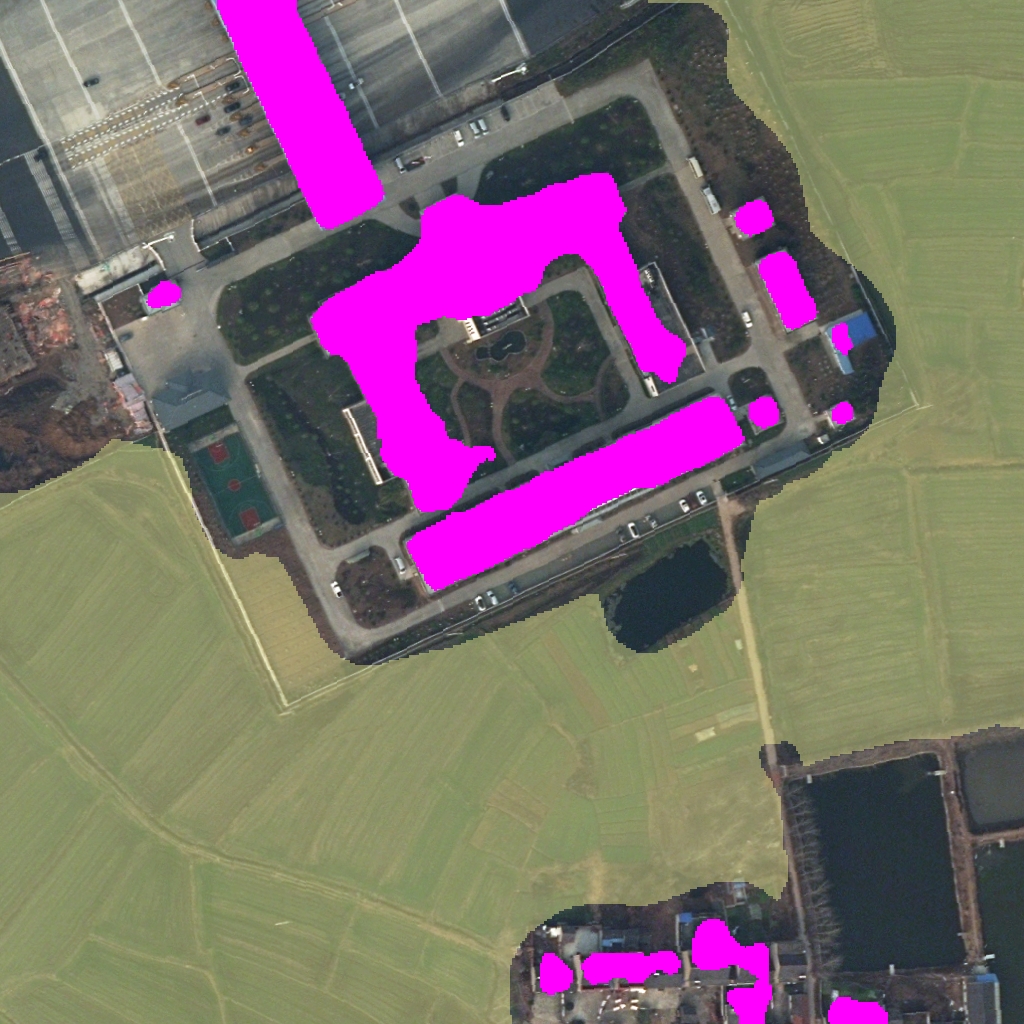}
\end{minipage}\hfill
\begin{minipage}[t]{0.58\linewidth}
\vspace{0pt}
\footnotesize\textbf{Dataset entry}
\begin{lstlisting}[basicstyle=\tiny\ttfamily, frame=single]
{ "id": "SQuID_1144",
  "image": "earthvqa_0.3m/2923.png",
  "question": "How many buildings (larger than
    0.01 hectares) are within 200m of
    agricultural land? (GSD: 0.3m)",
  "answer": 6,
  "type": "building_proximity",
  "tier": 2, "gsd": 0.3,
  "acceptable_range": [ 4, 8 ] }
\end{lstlisting}
\footnotesize\textbf{Program written by GPT-5}
\begin{lstlisting}[basicstyle=\tiny\ttfamily, frame=single]
gsd = 0.3
seg_result = segment_image_from_path(IMAGE_PATH,
    ["agric", "roof"], gsd=gsd)
shapes = seg_result["shapes"]
agric_shapes = [s for s in shapes
    if s["class_type"] == "agric"]
roofs = [s for s in shapes
    if s["class_type"] == "roof"]
min_hectares = 0.01
large_roofs = [s for s in roofs
    if s.get("area_hectares", 0) > min_hectares]
distance_m = 200.0
roofs_within = find_shapes_within_distance(
    large_roofs, agric_shapes,
    distance_m, resolution=gsd)
answer = len(roofs_within)
\end{lstlisting}
\footnotesize\textbf{Output}
\begin{lstlisting}[basicstyle=\tiny\ttfamily, frame=single]
answer = 7
\end{lstlisting}
\end{minipage}
\caption{\textbf{Full QASR pipeline example (GPT-5 + ConvNeXt).} Left, the image and the predicted masks. Right, the dataset entry, the generated program and its output. The answer 7 falls within the tolerance range [4, 8] of the reference answer 6.}
\label{fig:full_pipeline}
\end{figure}

All program generators receive the same developer prompt (\cref{fig:developer_prompt}). Its API descriptions are generated from the function docstrings and its topics from the segmentation models available on the server, so both can be extended without retraining. \Cref{fig:code_examples_supp} gives short programs for three question types, \cref{fig:full_pipeline} follows one question from the dataset entry to the final answer, and \cref{fig:gpt_oss_examples} shows a correct and an incorrect prediction from gpt-oss-120B programs.

\subsection{Additional Results}
\label{sec:additional_results}

\Cref{tab:full_results_by_type} gives accuracy by question type for 12 configurations. For Qwen3-VL-30B-A3B-Thinking, the answer is the text after the closing \texttt{</think>} token (token ID 151668), and \cref{fig:qwen_thinking_traces} shows two of its reasoning traces.

\subsection{Reference-Mask Replay}
\label{sec:gt_replay}

For GPT-5 and Llama-3.1-8B, we replay the logged programs on the 1,950 mask-derived questions in the same sandbox. The only change is the segmentation call, which returns shapes computed from the reference masks of the requested classes instead of predicted masks. The geometry functions, scoring rule and tolerance ranges are unchanged, and no program is regenerated. The 50 NAIP questions have human-consensus answers and no source masks, so they are excluded. \Cref{tab:results,tab:oracle} report the accuracies and outcome transitions. The replay mode is included in the released evaluation code.

\subsection{SQuID-Train and Fine-Tuning}
\label{sec:squid_train}

SQuID-Train is generated with the same pipeline, templates and quota-based sampler as the benchmark, on 1,443 labeled images of the source datasets that do not overlap with the benchmark images. It contains 75,687 question-answer pairs. We fine-tune Qwen2.5-VL-7B-Instruct~\cite{bai2025qwen25vl} on SQuID-Train with LoRA~\cite{hu2022lora} of rank 16 and a batch size of 2, and evaluate it on SQuID with the protocol of \cref{sec:experiments}. Its accuracy rises from 16.1\% to 40.7\%, with 53.5\%, 56.0\% and 13.2\% on Tiers~1 to~3.

\subsection{Human Reference}
\label{sec:human_ref}

We score each of the 500 annotator responses (50 NAIP questions, 10 annotators each) exactly as a model output is scored. Numeric answers must fall within the tolerance range and categorical answers must match exactly. Of the 500 responses, 313 are accepted (62.6\%). On the same 50 questions, 13 of the evaluated systems score between 12\% and 46\%. The same annotators' consensus defines the reference, so the aggregate answer matches it by construction. We report 62.6\% as individual-response agreement rather than an independent upper bound. The annotators were researchers trained on the interface, including its grid-cell selection and distance ruler, and we did not evaluate a separate cohort of RS experts.


\begin{table}[tp]
\centering
\caption{\textbf{ConvNeXt-UNet training configuration.} The same architecture is used for all four datasets.}
\label{tab:convnext_training}
\small
\begin{tabular}{ll}
\toprule
\textbf{Parameter} & \textbf{Value} \\
\midrule
Encoder & ConvNeXt, pretrained on ImageNet \\
Decoder & U-Net \\
Parameters & About 180M per model \\
Loss & Cross-entropy \\
Optimizer & Adam \\
Learning rate & $10^{-4}$ \\
Augmentation & Random affine patching and color augmentation \\
 & (RGB shifts, brightness, contrast, saturation, gamma) \\
Training data & EarthVQA, DeepGlobe, Solar, AIRS \\
Resolution & Trained at several GSDs per dataset \\
Instance segmentation & Same architecture with a background class \\
Post-processing & Per-class weighting and mode box filtering \\
\midrule
\multicolumn{2}{l}{\textit{Validation performance}} \\
EarthVQA & 81.2\% F1 (6 classes, semantic) \\
DeepGlobe & 80.0\% F1 (6 classes, semantic) \\
Solar & 97.0\% F1 (1 class, semantic) \\
AIRS & 98.4\% Dice (1 class, instance) \\
\bottomrule
\end{tabular}
\end{table}

\begin{table}[tp]
\centering
\caption{\textbf{Topics available to QASR with the ConvNeXt models.} When two models predict a class, the topic takes the maximum of their logits. The urban topic combines three source classes.}
\label{tab:QASR_classes}
\small
\begin{tabular}{lll}
\toprule
\textbf{Topic} & \textbf{Type} & \textbf{Source classes} \\
\midrule
urban & semantic & DeepGlobe urban, EarthVQA road, EarthVQA building \\
forest & semantic & DeepGlobe forest, EarthVQA forest \\
agric & semantic & DeepGlobe agriculture, EarthVQA agriculture \\
grass & semantic & DeepGlobe rangeland \\
barren & semantic & DeepGlobe barren, EarthVQA barren \\
water & semantic & DeepGlobe water, EarthVQA water \\
solar & semantic & Solar panels \\
roof & instance & AIRS roofs \\
\bottomrule
\end{tabular}
\end{table}

\begin{table}[tp]
\centering
\caption{\textbf{DINOv3 + Mask2Former configuration and results} on the EarthVQA validation split.}
\label{tab:dinov3_results}
\small
\setlength{\tabcolsep}{4pt}
\begin{tabular}{ll@{\hspace{2.5em}}lr}
\toprule
\multicolumn{2}{c}{\textbf{Training configuration}} & \multicolumn{2}{c}{\textbf{Results}} \\
\cmidrule(r{2.5em}){1-2}\cmidrule(l){3-4}
Backbone & DINOv3 ViT-L/16, frozen & Best mIoU & 57.58\% \\
Pretraining & SAT-493M, about 300M parameters & Pixel accuracy & 73.50\% \\
Decoder & Adapter and Mask2Former & Best epoch & 60 of 200 \\
 & (56.6M parameters) & Training time & 6.2 hours \\
Hidden dimension & 256 & & \\
Batch size & 20 (10 per GPU) & \multicolumn{2}{c}{\textit{IoU per class}} \\
Optimizer & AdamW, learning rate $10^{-4}$ & Water & 73.43\% \\
Loss & Class-weighted cross-entropy and Dice & Agriculture & 66.43\% \\
Precision & bfloat16 & Road & 63.72\% \\
Normalization & SAT-493M statistics$^{\dagger}$ & Building & 61.86\% \\
GPUs & 2 H100, data parallel & Playground & 59.42\% \\
 & & Background & 50.39\% \\
 & & Forest & 46.40\% \\
 & & Barren & 39.01\% \\
\bottomrule
\end{tabular}
\\[4pt]
{\scriptsize $^{\dagger}$Mean [0.430, 0.411, 0.296] and standard deviation [0.213, 0.156, 0.143].}
\end{table}

\begin{table}[tp]
\centering
\caption{\textbf{Cross-domain evaluation on GEOBench-VLM} (\texttt{Single} split, 3,211 five-way multiple-choice questions). For QASR, the tasks other than counting use GPT-5 end to end.}
\label{tab:geobench}
\small
\begin{tabular}{llccc}
\toprule
\textbf{Method} & \textbf{Segmentation} & \textbf{Counting} & \textbf{Other tasks} & \textbf{Overall} \\
\midrule
GPT-5, end to end & & \textbf{29.6\%} & 57.3\% & \textbf{48.2\%} \\
QASR with GPT-5 & SAM-3 & 15.8\% & 57.3\% & 43.7\% \\
QASR with GPT-5 & RemoteSAM & 15.6\% & 57.3\% & 43.6\% \\
\bottomrule
\end{tabular}

\vspace{10pt}
\begin{tabular}{lcc}
\toprule
\textbf{Counting task} & \textbf{QASR with RemoteSAM} & \textbf{GPT-5, end to end} \\
\midrule
Vehicles & \textbf{50.7\%} & 18.0\% \\
Buildings & 10.6\% & \textbf{33.5\%} \\
Aircraft & 13.0\% & \textbf{43.0\%} \\
\bottomrule
\end{tabular}
\end{table}

\begin{table}[tp]
\centering
\caption{\textbf{Classes in SQuID and their minimum area thresholds.} Questions state these thresholds to reduce the effect of small segmentation artifacts.}
\label{tab:classes_supp}
\small
\begin{tabular}{lc}
\toprule
\textbf{Class} & \textbf{Minimum area} \\
\midrule
\multicolumn{2}{l}{\textit{DeepGlobe, 0.5\,m GSD}} \\
Urban area & 0.1 ha \\
Vegetation$^{*}$ & 0.125 ha \\
Water bodies & 0.1 ha \\
Barren land & 0.125 ha \\
\midrule
\multicolumn{2}{l}{\textit{EarthVQA, 0.3\,m GSD}} \\
Buildings & 0.01 ha \\
Agricultural land & 0.125 ha \\
Forest area & 0.125 ha \\
Water bodies & 0.1 ha \\
Barren land & 0.125 ha \\
\midrule
\multicolumn{2}{l}{\textit{Solar, 0.3\,m GSD}} \\
Solar panels & 0.01 ha \\
\midrule
\multicolumn{2}{l}{\textit{NAIP, 1.0\,m GSD}} \\
Buildings & 0.01 ha \\
Urban area & 0.1 ha \\
Water bodies & 0.1 ha \\
Agricultural land & 0.125 ha \\
Forest area & 0.125 ha \\
Barren land & 0.125 ha \\
Grass and rangeland & 0.125 ha \\
\bottomrule
\end{tabular}
\\[4pt]
{\scriptsize $^{*}$In DeepGlobe, vegetation groups agriculture, rangeland and forest.}
\end{table}

\begin{table}[tp]
\centering
\caption{\textbf{Question examples from SQuID}, one per type. The GSD is appended to each question in the dataset, and tolerance ranges follow \cref{sec:dataset}.}
\label{tab:questions_supp}
\scriptsize
\setlength{\tabcolsep}{4pt}
\renewcommand{\arraystretch}{1.1}
\begin{tabular}{l>{\raggedright\arraybackslash}p{6.4cm}ccl}
\toprule
\textbf{Type} & \textbf{Question} & \textbf{GSD} & \textbf{Answer} & \textbf{Range} \\
\midrule
\multicolumn{5}{l}{\cellcolor{blue!10}\textit{Tier 1: Basic Quantification (710 questions, 35.5\%)}} \\
percentage & What percentage of the image is covered by barren land? & 0.3\,m & 23.36\% & [21.6, 25.1] \\
count & How many separate urban area regions are there? When counting, ignore patches smaller than 0.1 hectares. & 0.5\,m & 4 & [3, 5] \\
size & What percentage of the image is covered by the largest vegetation region, among regions larger than 0.125 hectares? & 0.5\,m & 75.81\% & [74.1, 77.6] \\
total\_area & What is the total area of solar panels in hectares, excluding installations smaller than 0.01 hectares? & 0.3\,m & 3.14 ha & [3.08, 3.2] \\
binary\_comparison & Is there more barren land than forest area in this image? & 0.3\,m & yes & exact \\
binary\_threshold & Is there more than 1 hectare of solar panels in this image, excluding installations smaller than 0.01 hectares? & 0.3\,m & no & exact \\
binary\_presence & Are there any solar panel installations larger than 0.01 hectares in this image? & 0.3\,m & yes & exact \\
binary\_multiple & Are there several separate solar installations larger than 0.01 hectares in this image? & 0.3\,m & yes & exact \\
\midrule
\multicolumn{5}{l}{\cellcolor{orange!10}\textit{Tier 2: Spatial Relationships (616 questions, 30.8\%)}} \\
proximity\_percentage & What percentage of the urban area is within 500m of vegetation? & 0.5\,m & 34.35\% & [32.1, 36.6] \\
proximity\_area & What is the total vegetation area (in hectares) within 200m of barren land? & 0.5\,m & 105.55 ha & [103.2, 107.9] \\
connectivity & How many separate agricultural land patches between 0.125 and 10 hectares are there? & 0.3\,m & 2 & [1, 3] \\
fragmentation & Is the forest area connected or fragmented, where fragmented means more than 5 patches larger than 0.125 hectares? & 0.3\,m & connected & exact \\
binary\_proximity & Is there any barren land within 100m of the urban area? & 0.5\,m & yes & exact \\
building\_proximity & How many buildings (larger than 0.01 hectares) are within 500m of agricultural land? & 0.3\,m & 4 & [3, 5] \\
building\_flood\_risk & How many buildings (larger than 0.01 hectares) are within 100m of water (flood risk)? & 0.3\,m & 0 & [0, 1] \\
building\_fire\_risk & How many buildings (larger than 0.01 hectares) are within 50m of forest (fire risk)? & 0.3\,m & 3 & [2, 4] \\
power\_calculation & What is the solar power potential of the panels in megawatts, assuming 200 W per square meter? & 0.3\,m & 0.74 MW & [0.72, 0.76] \\
\midrule
\multicolumn{5}{l}{\cellcolor{red!10}\textit{Tier 3: Complex Multi-Condition (674 questions, 33.7\%)}} \\
complex\_multi\_condition & Find barren land patches larger than 1 hectare, then calculate how much of their area (in hectares) lies within 200m of forest. & 0.3\,m & 1.49 ha & [1.46, 1.52] \\
complex\_veg\_water & Find vegetation patches larger than 2 hectares, then calculate how much of their area (in hectares) lies within 200m of water. & 0.5\,m & 63.93 ha & [62.5, 65.4] \\
complex\_agri\_water & Find agricultural land patches larger than 2 hectares, then calculate how much of their area (in hectares) lies within 200m of water. & 0.3\,m & 4.47 ha & [4.37, 4.57] \\
complex\_urban\_fire & Find urban areas larger than 1 hectare, then calculate how much of their area (in hectares) lies within 50m of vegetation. & 0.5\,m & 3.02 ha & [2.95, 3.09] \\
complex\_urban\_flood & Find urban areas larger than 1 hectare, then calculate how much of their area (in hectares) lies within 100m of water. & 0.5\,m & 4.03 ha & [3.94, 4.12] \\
complex\_size\_filter & What is the total area (in hectares) of utility-scale solar installations, that is, installations larger than 5 hectares? & 0.3\,m & 0 ha & [0.0, 0.0] \\
complex\_average & What is the average size (in hectares) of solar installations, excluding installations smaller than 0.01 hectares? & 0.3\,m & 1.33 ha & [1.31, 1.35] \\
\bottomrule
\end{tabular}
\end{table}

\begin{table}[tp]
\centering
\caption{\textbf{The 24 question types in SQuID.}}
\label{tab:types_supp}
\small
\setlength{\tabcolsep}{4pt}
\begin{tabular}{lrrl}
\toprule
\textbf{Question type} & \textbf{Count} & \textbf{Share} & \textbf{Description} \\
\midrule
\multicolumn{4}{l}{\textit{Tier 1: Basic Quantification (710 questions, 35.5\%)}} \\
count & 178 & 8.9\% & Number of regions above an area threshold \\
binary\_comparison & 172 & 8.6\% & Whether one class covers more area than another \\
size & 166 & 8.3\% & Size of the largest or smallest region \\
percentage & 157 & 7.8\% & Percentage of the image covered by one class \\
binary\_threshold & 11 & 0.6\% & Whether the total area exceeds a threshold \\
binary\_presence & 10 & 0.5\% & Whether any instance exceeds a size \\
binary\_multiple & 10 & 0.5\% & Whether several instances exceed a size \\
total\_area & 6 & 0.3\% & Total area of all instances \\
\midrule
\multicolumn{4}{l}{\textit{Tier 2: Spatial Relationships (616 questions, 30.8\%)}} \\
proximity\_percentage & 123 & 6.2\% & Percentage of class A within distance D of class B \\
binary\_proximity & 122 & 6.1\% & Whether class A lies within distance D of class B \\
proximity\_area & 107 & 5.4\% & Area of class A within distance D of class B \\
connectivity & 104 & 5.2\% & Number of regions within a size range \\
fragmentation & 98 & 4.9\% & Whether a class is connected or fragmented \\
building\_proximity & 35 & 1.8\% & Buildings within distance D of a land cover \\
power\_calculation & 14 & 0.7\% & Solar capacity in megawatts from panel area \\
building\_fire\_risk & 9 & 0.5\% & Buildings within 50\,m of forest \\
building\_flood\_risk & 4 & 0.2\% & Buildings within 100\,m of water \\
\midrule
\multicolumn{4}{l}{\textit{Tier 3: Complex Multi-Condition (674 questions, 33.7\%)}} \\
complex\_multi\_condition & 490 & 24.5\% & Size filter, proximity and area computation \\
complex\_agri\_water & 81 & 4.1\% & Agricultural patches over 2\,ha within 200\,m of water \\
complex\_veg\_water & 32 & 1.6\% & Vegetation patches over 2\,ha within 200\,m of water \\
complex\_urban\_fire & 32 & 1.6\% & Urban areas over 1\,ha within 50\,m of vegetation \\
complex\_urban\_flood & 18 & 0.9\% & Urban areas over 1\,ha within 100\,m of water \\
complex\_average & 15 & 0.8\% & Average size, excluding small artifacts \\
complex\_size\_filter & 6 & 0.3\% & Total area above a utility-scale threshold \\
\midrule
\textbf{Total} & \textbf{2,000} & \textbf{100\%} & \\
\bottomrule
\end{tabular}
\\[4pt]
{\scriptsize EarthVQA (1,241), DeepGlobe (612), Solar (97), NAIP (50). GSD 0.3\,m (1,338), 0.5\,m (612), 1.0\,m (50).}
\end{table}

\begin{figure}[tp]
\centering
\begin{tcolorbox}[colback=gray!5, colframe=black!70, width=0.98\textwidth, arc=1mm, boxrule=0.5pt, title={\textbf{QASR Developer Prompt (identical for all program generators)}}, top=2mm, bottom=2mm]
\footnotesize
You are a code generator for geospatial image analysis.

\textbf{CRITICAL RULES:}
\begin{itemize}[leftmargin=*,nosep,topsep=1pt,itemsep=1pt]
    \item Output \textbf{ONLY} executable Python code
    \item \textbf{NO} explanatory text before or after code
    \item \textbf{NO} imports (everything is already imported)
    \item \textbf{NO} markdown code blocks
    \item Store final result in variable {\color{blue!70!black}\texttt{answer}}
\end{itemize}

\textbf{Available APIs (already imported, do not import anything):}

{\color{blue!70!black}\textbf{\texttt{segment\_image\_from\_path(image\_input, topics, min\_area\_pixels=0, gsd=1.0)}}}

Load and segment an image from a file path or PIL Image using mask-based approach.
\textbf{Args:} \texttt{image\_input} (str | PIL.Image), \texttt{topics} (list[str]), \texttt{min\_area\_pixels} (int, optional), \texttt{gsd} (float, optional).
\textbf{Returns:} dict with \texttt{shapes} (list of shape dicts containing \texttt{id}, \texttt{class\_type}, \texttt{area\_pixels}, \texttt{area\_hectares}, \texttt{polygon}, etc.), \texttt{image\_width}, \texttt{image\_height}, \texttt{total\_pixels}

{\color{blue!70!black}\textbf{\texttt{find\_shapes\_within\_distance(target, reference, distance\_m, resolution)}}}

\textbf{CUTS/CLIPS} target shapes to portions within distance\_m of any reference shape.
\textbf{Returns:} list[dict] of NEW clipped shape dicts.

{\color{blue!70!black}\textbf{\texttt{calculate\_shape\_distances(target, reference, resolution)}}}

\textbf{MODIFIES} target shapes in place, adding \texttt{distance\_meters} field. \textbf{Returns:} same list.

\textbf{Topics:} {\color{blue!70!black}\texttt{urban}}, {\color{blue!70!black}\texttt{forest}}, {\color{blue!70!black}\texttt{agric}}, {\color{blue!70!black}\texttt{grass}}, {\color{blue!70!black}\texttt{barren}}, {\color{blue!70!black}\texttt{water}}, {\color{blue!70!black}\texttt{solar}}, {\color{blue!70!black}\texttt{roof}}

\textbf{Context:} {\color{blue!70!black}\texttt{IMAGE\_PATH}} contains the image path. Do not open with PIL.

\textbf{Question:} \textit{[User's geospatial question with GSD]}. Generate ONLY the Python code.
\end{tcolorbox}
\caption{\textbf{QASR developer prompt.} The API descriptions are generated from the function docstrings and the topics from the segmentation models available on the server, so both can be extended without retraining.}
\label{fig:developer_prompt}
\end{figure}

\begin{figure}[tp]
\centering
\begin{tcolorbox}[colback=gray!5, colframe=black!50, width=0.98\textwidth, arc=1mm, boxrule=0.5pt, title={\textbf{Further QASR Program Examples}}, top=2mm, bottom=2mm]
\footnotesize
\textbf{Percentage.} ``What percentage of the image is covered by forest?''
\begin{lstlisting}[language=Python, basicstyle=\tiny\ttfamily, frame=single]
result = segment_image_from_path(IMAGE_PATH, ["forest"], gsd=0.5)
forest_px = sum(s["area_pixels"] for s in result["shapes"] if s["class_type"] == "forest")
answer = (forest_px / result["total_pixels"]) * 100
\end{lstlisting}
\textbf{Counting with a size filter.} ``How many separate agricultural land regions larger than 0.125 hectares are there?''
\begin{lstlisting}[language=Python, basicstyle=\tiny\ttfamily, frame=single]
result = segment_image_from_path(IMAGE_PATH, ["agric"], gsd=0.3)
answer = len([s for s in result["shapes"] if s["area_hectares"] > 0.125])
\end{lstlisting}
\textbf{Proximity.} ``What percentage of the urban area is within 100m of water?''
\begin{lstlisting}[language=Python, basicstyle=\tiny\ttfamily, frame=single]
result = segment_image_from_path(IMAGE_PATH, ["urban", "water"], gsd=0.5)
urban = [s for s in result["shapes"] if s["class_type"] == "urban"]
water = [s for s in result["shapes"] if s["class_type"] == "water"]
near = find_shapes_within_distance(urban, water, 100.0, 0.5)
answer = sum(s["area_pixels"] for s in near) / sum(s["area_pixels"] for s in urban) * 100
\end{lstlisting}
\end{tcolorbox}
\caption{\textbf{Further QASR program examples} for a percentage, a counting and a proximity question. \Cref{sec:QASR} shows a multi-condition example and \cref{fig:full_pipeline} a full pipeline.}
\label{fig:code_examples_supp}
\end{figure}

\begin{figure}[tp]
\centering
\begin{minipage}[t]{0.48\textwidth}
\vspace{0pt}
\begin{tcolorbox}[colback=green!5, colframe=green!60!black, title={\textbf{$\checkmark$ Correct: Percentage (SQuID\_1378)}}, fonttitle=\small\bfseries, top=1.5mm, bottom=1.5mm, equal height group=gptoss]
\scriptsize
\textbf{Question.} What percentage of the image is covered by agricultural land? (GSD: 0.3m)

\textbf{Reference.} 14.6\%, range [12.86, 16.34]

\textbf{Program.}
\begin{lstlisting}[basicstyle=\tiny\ttfamily,breaklines=true]
result = segment_image_from_path(IMAGE_PATH,
  ["urban","forest","agric","grass",
   "barren","water","solar","roof"], gsd=gsd)
shapes = result["shapes"]
total_pixels = result["total_pixels"]
topic_counts = {}
for shape in shapes:
    t = shape["class_type"]
    topic_counts[t] = topic_counts.get(t, 0) + 1
print("Initial shape counts:", topic_counts)
agric_pixels = sum(s["area_pixels"] for s in shapes
    if s["class_type"] == "agric")
percentage = (agric_pixels / total_pixels) * 100
answer = percentage
\end{lstlisting}
\textbf{Output.}
\begin{lstlisting}[basicstyle=\tiny\ttfamily,breaklines=true]
Initial shape counts: {'urban': 28, 'forest': 34,
'agric': 7, 'grass': 34, 'barren': 13, 'water': 3,
'solar': 1, 'roof': 5}
\end{lstlisting}
{\color{green!70!black}\textbf{Prediction 13.66\% $\rightarrow$ correct}}
\end{tcolorbox}
\end{minipage}
\hfill
\begin{minipage}[t]{0.48\textwidth}
\vspace{0pt}
\begin{tcolorbox}[colback=red!5, colframe=red!60!black, title={\textbf{$\times$ Incorrect: Count (SQuID\_1374)}}, fonttitle=\small\bfseries, top=1.5mm, bottom=1.5mm, equal height group=gptoss]
\scriptsize
\textbf{Question.} How many separate urban area regions are there? When counting, ignore patches smaller than 0.1 hectares. (GSD: 0.5m)

\textbf{Reference.} 5, range [4, 6]

\textbf{Program.}
\begin{lstlisting}[basicstyle=\tiny\ttfamily,breaklines=true]
result = segment_image_from_path(IMAGE_PATH,
  ["urban","forest","agric","grass",
   "barren","water","solar","roof"], gsd=gsd)
shapes = result["shapes"]
for t in ["urban","forest","agric","grass",
          "barren","water","solar","roof"]:
    count = sum(1 for s in shapes
        if s["class_type"] == t)
    print(f"{t}: {count}")
urban_large = [s for s in shapes
    if s["class_type"] == "urban"
    and s.get("area_hectares", 0) >= 0.1]
answer = len(urban_large)
\end{lstlisting}
\textbf{Output.}
\begin{lstlisting}[basicstyle=\tiny\ttfamily]
urban: 23, forest: 7, agric: 19, grass: 0
barren: 8, water: 5, solar: 1, roof: 9
\end{lstlisting}
{\color{red!70!black}\textbf{Prediction 9 $\rightarrow$ incorrect}}

\textit{Nine predicted urban regions pass the area threshold, compared with a reference count of 5.}
\end{tcolorbox}
\end{minipage}
\caption{\textbf{gpt-oss-120B program examples.} Left, a percentage prediction within the tolerance range. Right, a count of 9, outside the reference range [4, 6].}
\label{fig:gpt_oss_examples}
\end{figure}

\begin{table}[tp]
\centering
\caption{\textbf{Accuracy (\%) by question type for 12 configurations.} QASR-A to QASR-F are the configurations of \cref{tab:results}, run with predicted masks. Rows follow \cref{tab:types_supp}, and the best value in each row is in bold.}
\label{tab:full_results_by_type}
\footnotesize
\setlength{\tabcolsep}{2.4pt}
\begin{tabular}{@{}lr cccccc cccccc@{}}
\toprule
& & \multicolumn{6}{c}{\textbf{QASR, predicted masks}} & \multicolumn{6}{c}{\textbf{End-to-end VLMs}} \\
\cmidrule(lr){3-8} \cmidrule(l){9-14}
\textbf{Question type} & \textbf{N} & \rotatebox{90}{QASR-A} & \rotatebox{90}{QASR-B} & \rotatebox{90}{QASR-C} & \rotatebox{90}{QASR-D} & \rotatebox{90}{QASR-E} & \rotatebox{90}{QASR-F} & \rotatebox{90}{GPT-5} & \rotatebox{90}{Qwen3-VL-30B-A3B} & \rotatebox{90}{Qwen3-VL-8B} & \rotatebox{90}{GeoChat} & \rotatebox{90}{Geo-R1} & \rotatebox{90}{Qwen2.5-VL-7B} \\
\midrule
\multicolumn{14}{@{}l}{\textit{Tier 1: Basic Quantification}} \\
count & 178 & \textbf{56.7} & 36.0 & 55.1 & 36.5 & 27.5 & 18.5 & 36.5 & 45.5 & 15.7 & 27.0 & 42.1 & 3.4 \\
binary\_comparison & 172 & 78.5 & 76.7 & 79.7 & 57.6 & 46.5 & 61.6 & 67.4 & 78.5 & 75.0 & \textbf{100} & 52.9 & 57.0 \\
size & 166 & \textbf{33.7} & 23.5 & \textbf{33.7} & 24.1 & 4.2 & 19.9 & 16.3 & 12.7 & 3.0 & 4.8 & 10.8 & 11.4 \\
percentage & 157 & \textbf{36.9} & 33.1 & 26.1 & 31.2 & 2.5 & 5.1 & 26.8 & 11.5 & 7.0 & 2.5 & 6.4 & 7.6 \\
binary\_threshold & 11 & \textbf{100} & \textbf{100} & \textbf{100} & 90.9 & 72.7 & 45.5 & \textbf{100} & \textbf{100} & \textbf{100} & 90.9 & \textbf{100} & \textbf{100} \\
binary\_presence & 10 & \textbf{100} & \textbf{100} & \textbf{100} & \textbf{100} & 90.0 & 30.0 & 90.0 & 50.0 & 90.0 & 90.0 & 90.0 & 90.0 \\
binary\_multiple & 10 & 90.0 & \textbf{100} & 90.0 & \textbf{100} & \textbf{100} & 80.0 & 80.0 & 60.0 & \textbf{100} & 40.0 & 90.0 & 70.0 \\
total\_area & 6 & 0.0 & 0.0 & 0.0 & 0.0 & 0.0 & 0.0 & \textbf{16.7} & 0.0 & 0.0 & 0.0 & 0.0 & 0.0 \\
\midrule
\multicolumn{14}{@{}l}{\textit{Tier 2: Spatial Relationships}} \\
proximity\_percentage & 123 & \textbf{40.6} & 37.4 & \textbf{40.6} & 34.1 & 15.4 & 35.8 & 19.5 & 26.0 & 24.4 & 13.0 & 13.0 & 10.6 \\
binary\_proximity & 122 & 77.9 & 86.9 & 68.0 & 86.1 & 72.1 & 23.8 & 80.3 & 81.2 & 84.4 & \textbf{91.0} & 63.9 & 56.6 \\
proximity\_area & 107 & \textbf{6.5} & 0.0 & 1.9 & 0.0 & 0.0 & 1.9 & 0.0 & 0.0 & 0.0 & 0.0 & 1.9 & 0.0 \\
connectivity & 104 & \textbf{74.0} & 51.0 & 63.5 & 49.0 & 43.3 & 29.8 & 37.5 & 44.2 & 9.6 & 53.9 & 33.6 & 5.8 \\
fragmentation & 98 & \textbf{81.6} & 63.3 & 69.4 & 50.0 & 66.3 & 70.4 & 26.5 & 26.5 & 74.5 & 22.4 & 21.4 & 66.3 \\
building\_proximity & 35 & 51.4 & 31.4 & 48.6 & 22.9 & 45.7 & 45.7 & 51.4 & 51.4 & 20.0 & 51.4 & \textbf{57.1} & 8.6 \\
power\_calculation & 14 & \textbf{7.1} & 0.0 & \textbf{7.1} & 0.0 & 0.0 & 0.0 & 0.0 & 0.0 & 0.0 & 0.0 & 0.0 & 0.0 \\
building\_fire\_risk & 9 & 33.3 & 33.3 & 33.3 & 22.2 & 33.3 & 33.3 & 22.2 & \textbf{44.4} & 33.3 & 33.3 & 33.3 & 11.1 \\
building\_flood\_risk & 4 & 50.0 & 50.0 & 50.0 & 25.0 & 50.0 & \textbf{75.0} & \textbf{75.0} & 50.0 & 25.0 & 50.0 & 25.0 & 25.0 \\
\midrule
\multicolumn{14}{@{}l}{\textit{Tier 3: Complex Multi-Condition}} \\
complex\_multi\_condition & 490 & \textbf{20.0} & 6.9 & 13.7 & 7.1 & 6.7 & 12.4 & 11.8 & 4.5 & 12.0 & 0.2 & 1.8 & 0.4 \\
complex\_agri\_water & 81 & \textbf{23.5} & 1.2 & 7.4 & 1.2 & 1.2 & 3.7 & 3.7 & 0.0 & 3.7 & 0.0 & 0.0 & 0.0 \\
complex\_veg\_water & 32 & 0.0 & 0.0 & 0.0 & 0.0 & 0.0 & 0.0 & 0.0 & 0.0 & 0.0 & 0.0 & \textbf{3.1} & 0.0 \\
complex\_urban\_fire & 32 & 12.5 & 9.4 & \textbf{15.6} & 6.2 & 6.2 & 9.4 & \textbf{15.6} & 6.2 & 12.5 & 3.1 & 0.0 & 0.0 \\
complex\_urban\_flood & 18 & 11.1 & 11.1 & 11.1 & 0.0 & \textbf{16.7} & \textbf{16.7} & \textbf{16.7} & 5.6 & 11.1 & 0.0 & 0.0 & 0.0 \\
complex\_average & 15 & 0.0 & \textbf{6.7} & 0.0 & \textbf{6.7} & 0.0 & \textbf{6.7} & 0.0 & 0.0 & 0.0 & 0.0 & 0.0 & 0.0 \\
complex\_size\_filter & 6 & 66.7 & 0.0 & 66.7 & 0.0 & 33.3 & \textbf{83.3} & 66.7 & 0.0 & 0.0 & 0.0 & 0.0 & 0.0 \\
\midrule
\textbf{Total} & \textbf{2000} & \textbf{42.0} & 32.1 & 36.9 & 29.0 & 22.3 & 23.4 & 28.1 & 26.4 & 24.9 & 24.2 & 20.4 & 16.1 \\
\bottomrule
\end{tabular}
\end{table}

\begin{figure}[tp]
\centering
\begin{minipage}[t]{0.48\textwidth}
\vspace{0pt}
\begin{tcolorbox}[colback=green!5, colframe=green!70!black, width=\textwidth, arc=1mm, title={\textbf{$\checkmark$ Correct: Proximity (SQuID\_0001)}}, fonttitle=\small\bfseries, top=1.5mm, bottom=1.5mm, equal height group=qwen]
\scriptsize
\textbf{Question.} What percentage of the urban area is within 500m of vegetation? (GSD: 0.5m)

\textbf{Reference.} 34.35\%, range [32.1, 36.6]

\textbf{Thinking trace, shortened.}
\begin{lstlisting}[language={},basicstyle=\tiny\ttfamily,backgroundcolor=\color{blue!5}]
<think>
The question asks for percentage of urban
area within 500m of vegetation. Image shows
satellite view with GSD 0.5m. For 500m
buffer: 1000 pixels. City with dense
buildings and scattered green areas.
Estimating: roughly 30-40%. Going with 35%.
</think>
\end{lstlisting}
\textbf{Answer.} \texttt{35}

{\color{green!70!black}\textbf{Result 35\% $\rightarrow$ correct}}
\end{tcolorbox}
\end{minipage}
\hfill
\begin{minipage}[t]{0.48\textwidth}
\vspace{0pt}
\begin{tcolorbox}[colback=red!5, colframe=red!70!black, width=\textwidth, arc=1mm, title={\textbf{$\times$ Incorrect: Area (SQuID\_0000)}}, fonttitle=\small\bfseries, top=1.5mm, bottom=1.5mm, equal height group=qwen]
\scriptsize
\textbf{Question.} What is the total vegetation area (in hectares) within 200m of barren land? (GSD: 0.5m)

\textbf{Reference.} 105.55 hectares, range [103.18, 107.92]

\textbf{Thinking trace, shortened.}
\begin{lstlisting}[language={},basicstyle=\tiny\ttfamily,backgroundcolor=\color{blue!5}]
<think>
Vegetation area within 200m of barren
land, GSD 0.5m. Need vegetation within
200m buffer. Without image processing
capability, this seems impossible...
Maybe answer is 20000? Going with 20000.
</think>
\end{lstlisting}
\textbf{Answer.} \texttt{20000}

{\color{red!70!black}\textbf{Result 20000 hectares $\rightarrow$ incorrect}}

\textit{The displayed trace gives an estimate without showing a buffer-area calculation.}
\end{tcolorbox}
\end{minipage}
\caption{\textbf{Qwen3-VL-30B-A3B-Thinking examples.} Left, a proximity estimate within the tolerance range. Right, an incorrect area estimate. The displayed traces are shortened.}
\label{fig:qwen_thinking_traces}
\end{figure}

\clearpage


\end{document}